%% file: main.tex
\documentclass{style/hcrg-nasa}

\issuer{Harvard Computational Robotics Group}
\edition{Preprint}
\logofile{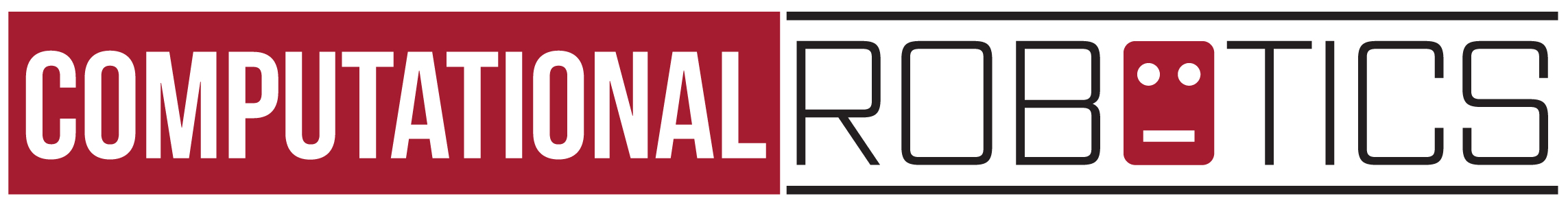}

\input{preamble_packages}
\input{preamble_symbols}

\input{shortcuts}
\input{preamble_formatting}

\title{\XMFD: 4D Reconstruction through Certifiable Geometric Optimization}
\author[1]{Haoyu Han}
\author[1]{Jack Benarroch Jedlicki}
\author[2]{Jiahui Lei}
\author[1]{Heng Yang}
\affiliation[1]{Harvard University}
\affiliation[2]{UC Berkeley}

\hypersetup{
  hypertexnames=false,
  pdftitle={XM4D: 4D Reconstruction through Certifiable Geometric Optimization},
  pdfauthor={Haoyu Han, Jack Benarroch Jedlicki, Jiahui Lei, Heng Yang},
  pdfkeywords={4D reconstruction, CUDA, embedded deformation graph, semidefinite relaxation, structure from motion}
}
\input{sections/abstract}
\website{https://computationalrobotics.seas.harvard.edu/XM4D/}
  {\texttt{https://computationalrobotics.seas.harvard.edu/XM4D/}}

\begin{document}
\maketitle

\par\vspace{4pt}
\noindent\begin{minipage}{\linewidth}
  \centering
  \includegraphics[width=\linewidth]{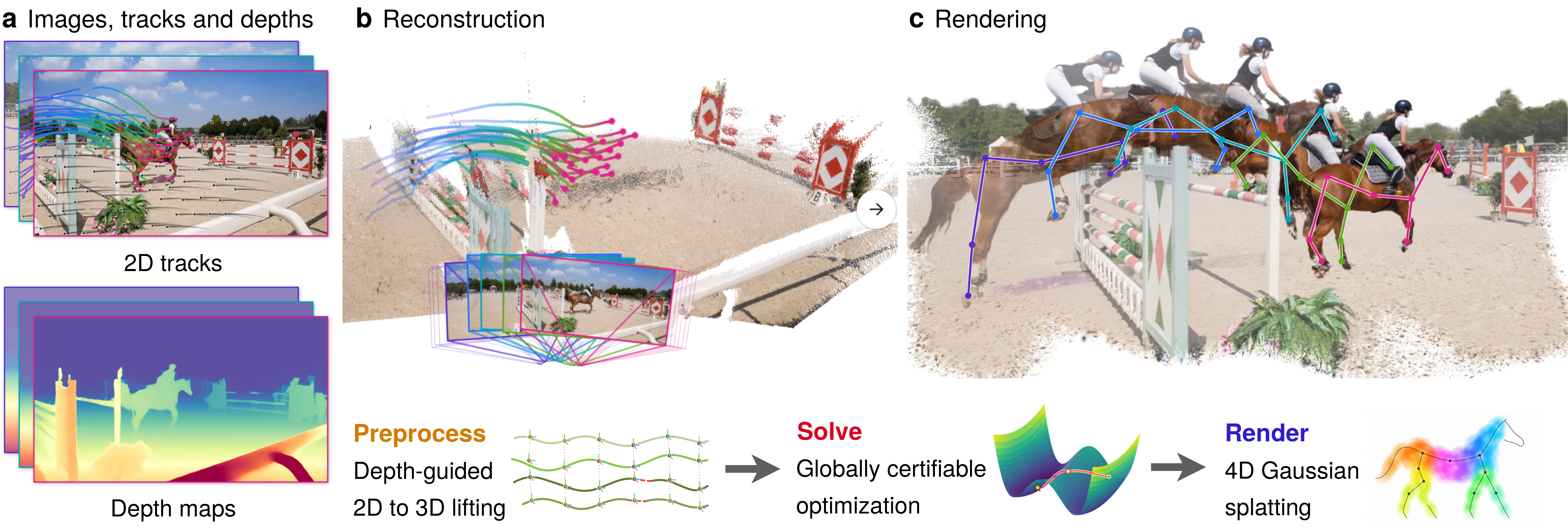}
  \captionsetup{skip=4pt,hypcap=false}
  \captionof{figure}{\textbf{\XMFD overview.} From (a)~2D tracks and monocular depth, a solver with certified global optimality followed by refinement recovers (b)~the cameras, the static geometry (points), and the 3D trajectories of dynamic points (colored curves), which initialize (c)~4D Gaussian splatting for rendering. The skeleton in (c) is a subsampled embedded deformation graph (EDG).}
  \label{fig:teaser}
\end{minipage}
\par\clearpage

\tcbset{
  enhanced, sharp corners, frame hidden,
  colback=crimson!5!white, borderline west={2pt}{0pt}{crimson},
  boxrule=0pt, boxsep=0pt,
  left=10pt, right=10pt, top=5pt, bottom=5pt,
  before skip=8pt, after skip=10pt
}

\input{sections/introduction}

\input{sections/related-work}
\input{sections/static}
\input{sections/dynamic}

\input{sections/pipeline}
\input{sections/experiments}

\FloatBarrier
\input{sections/conclusion}
\input{sections/acknowledgements}

\setlength{\bibsep}{1pt plus 1pt minus 1pt}
\bibliographystyle{style/plainnat-crg}
\bibliography{bibliography/myRefs,bibliography/refs}

\clearpage
\beginappendix
\begingroup
\small
\input{sections/app-tables}

\input{sections/app-proofs}

\input{sections/app-subopt}

\input{sections/app-datasets}

\input{sections/app-implementation}
\endgroup
\end{document}

%% file: preamble_packages.tex
\usepackage[utf8]{inputenc}
\usepackage{comment}
\usepackage{relsize}
\usepackage{ifthen}
\usepackage[colorinlistoftodos]{todonotes}
\usepackage{subfig}
\usepackage{textcomp}
\usepackage{verbatim}
\usepackage{rotating}
\usepackage{psfrag}
\usepackage{epsfig}
\usepackage{latexsym}
\usepackage{xstring}
\usepackage{prettyref}
\usepackage{bigdelim}
\usepackage{listings}
\usepackage{tikz}
\usetikzlibrary{matrix,calc}

\graphicspath{{figures/}}
\setlist[itemize]{topsep=4pt,itemsep=2pt,parsep=0pt}
\SetAlgoVlined
\LinesNumbered

%% file: preamble_symbols.tex
\newrefformat{prob}{Problem\,\ref{#1}}
\newrefformat{def}{Definition\,\ref{#1}}
\newrefformat{sec}{Section\,\ref{#1}}
\newrefformat{sub}{Section\,\ref{#1}}
\newrefformat{prop}{Proposition\,\ref{#1}}
\newrefformat{app}{Appendix\,\ref{#1}}
\newrefformat{alg}{Algorithm\,\ref{#1}}
\newrefformat{cor}{Corollary\,\ref{#1}}
\newrefformat{thm}{Theorem\,\ref{#1}}
\newrefformat{lem}{Lemma\,\ref{#1}}
\newrefformat{fig}{Fig.\,\ref{#1}}
\newrefformat{tab}{Table\,\ref{#1}}

\newtheorem{theorem}{Theorem}

\newtheorem{proposition}[theorem]{Proposition}

\newcommand{\bdmath}{\begin{dmath}}
\newcommand{\edmath}{\end{dmath}}
\newcommand{\beq}{\begin{equation}}
\newcommand{\eeq}{\end{equation}}
\newcommand{\bdm}{\begin{displaymath}}
\newcommand{\edm}{\end{displaymath}}
\newcommand{\bea}{\begin{eqnarray}}
\newcommand{\eea}{\end{eqnarray}}
\newcommand{\beal}{\beq \begin{array}{ll}}
\newcommand{\eeal}{\end{array} \eeq}
\newcommand{\beas}{\begin{eqnarray*}}
\newcommand{\eeas}{\end{eqnarray*}}
\newcommand{\ba}{\begin{array}}
\newcommand{\ea}{\end{array}}
\newcommand{\bit}{\begin{itemize}}
\newcommand{\eit}{\end{itemize}}
\newcommand{\ben}{\begin{enumerate}}
\newcommand{\een}{\end{enumerate}}

\renewcommand{\boldsymbol}[1]{{\bm #1}}

\newcommand{\hide}[1]{}

\newcommand{\hiddenText}{{\color{gray} hidden text.}}
\newcommand{\hideWithText}[1]{\hiddenText}

\newcommand{\norm}[1]{\left\| #1 \right\|}

\newcommand{\blue}[1]{{\color{blue}#1}}

\newcommand{\linkToPdf}[1]{\href{#1}{\blue{(pdf)}}}
\newcommand{\linkToPpt}[1]{\href{#1}{\blue{(ppt)}}}
\newcommand{\linkToCode}[1]{\href{#1}{\blue{(code)}}}
\newcommand{\linkToWeb}[1]{\href{#1}{\blue{(web)}}}
\newcommand{\linkToVideo}[1]{\href{#1}{\blue{(video)}}}
\newcommand{\linkToMedia}[1]{\href{#1}{\blue{(media)}}}
\newcommand{\award}[1]{\xspace} 



%% file: shortcuts.tex
\renewcommand{\norm}[1]{\left\lVert #1 \right\rVert}

\newcommand{\bmat}{\left[ \begin{array}}
\newcommand{\emat}{\end{array}\right]}

\definecolor{xmblue}{RGB}{31,78,140}
\definecolor{xmorange}{RGB}{230,120,30}
\newcommand{\XMFD}{\textbf{\textsc{\textcolor{xmblue}{XM}\textcolor{xmorange}{4D}}}\xspace}
\newcommand{\XM}{\textsc{XM}\xspace}

\newcommand{\oom}{\textcolor{red}{OOM}}
\newcommand{\timeout}{\textcolor{red}{$>$1h}}
\newcommand{\err}{\textcolor{red}{err}}

%% file: preamble_formatting.tex
\tcbset{crgstatement/.style={
  enhanced, sharp corners, frame hidden,
  colback=crimson!5!white, borderline west={2pt}{0pt}{crimson},
  boxrule=0pt, boxsep=0pt,
  left=10pt, right=10pt, top=5pt, bottom=5pt,
  before skip=8pt, after skip=10pt
}}
\tcolorboxenvironment{theorem}{crgstatement}
\tcolorboxenvironment{proposition}{crgstatement}
\tcolorboxenvironment{lemma}{crgstatement}
\tcolorboxenvironment{corollary}{crgstatement}
\tcolorboxenvironment{definition}{crgstatement}
\tcolorboxenvironment{assumption}{crgstatement}
\tcolorboxenvironment{remark}{crgstatement}
\tcolorboxenvironment{example}{crgstatement}
\tcolorboxenvironment{conjecture}{crgstatement}
\tcolorboxenvironment{problem}{crgstatement}
\tcolorboxenvironment{trule}{crgstatement}

\renewcommand{\linkToPdf}[1]{}
\renewcommand{\linkToPpt}[1]{}
\renewcommand{\linkToCode}[1]{}
\renewcommand{\linkToWeb}[1]{}
\renewcommand{\linkToVideo}[1]{}
\renewcommand{\linkToMedia}[1]{}

\renewcommand*{\backref}[1]{}

\renewcommand*{\backrefalt}[4]{%
  \ifnum #1>0\relax\space#2\fi
}

%% file: sections/abstract.tex
\abstract{
Accurate 4D reconstruction remains challenging: jointly estimating camera motion, scene geometry, and non-rigid deformation entails highly nonconvex optimization, making global optimality difficult to guarantee. We present \XMFD, a framework for accurate and efficient 4D reconstruction built upon \emph{certifiably globally optimal} geometric optimization. By separating the estimation of camera motion from non-rigid deformation, we formulate both stages as structured quadratically constrained quadratic programs (QCQP) that admit semidefinite programming (SDP) relaxations and verifiable global optimality certificates. For camera motion estimation, we lift static point tracks into 3D using foundational depth prediction models and formulate a scaled bundle adjustment problem that jointly estimates camera poses, depth scales, and static landmarks. By analytically eliminating translations and landmarks, we reduce the problem to a QCQP over scaled rotations. For non-rigid reconstruction, we represent dynamic motion using an embedded deformation graph (EDG) and show that its tracking, rigidity, and temporal regularization objectives similarly reduce to a rotation-only QCQP. Both problems admit SDP relaxations that we solve efficiently with a GPU-accelerated Burer--Monteiro solver based on Riemannian optimization. The recovered geometry further initializes dynamic Gaussian splatting for photometric refinement. Extensive experiments across seven benchmarks demonstrate accurate camera motion estimation on both static and dynamic scenes, efficient optimization that scales to thousands of frames, and state-of-the-art novel-view rendering quality.
}
\keywords{
4D reconstruction, embedded deformation graph, semidefinite relaxation, structure from motion
}

%% file: sections/introduction.tex
\section{Introduction}\label{sec:intro}

4D reconstruction aims to recover the time-varying 3D structure of a scene from visual observations. It supports applications including dynamic novel-view rendering~\cite{li21cvpr-neuralsceneflow,lombardi19tog-neuralvolumes}, deformable and articulated object reconstruction~\cite{newcombe15cvpr-dynamicfusion,innmann16eccv-volumedeform,yang21cvpr-lasr}, digital humans and digital twins~\cite{peng21cvpr-neuralbody,weng22cvpr-humannerf,jiang25iccv-phystwin}, and motion retargeting for humanoid control~\cite{allshire25corl-retarget,he24corl-omnih2o,liu26arxiv-grasp}. Despite its broad utility, reconstructing a dynamic scene from monocular images or videos remains highly challenging~\cite{gao21iccv-dynamicsynthesis,gao22neurips-dycheck}. Camera and object motion are entangled in the visual observations~\cite{chen25iccv-backontrack}, while dynamic geometry itself is weakly constrained and may undergo complex non-rigid deformation. Long sequences introduce further difficulties, including fragmented point tracks, noisy and scale-inconsistent monocular depth, and temporally incomplete observations of the dynamic scene~\cite{lei25cvpr-mosca,wang25iccv-shapeofmotion}.

Recent 4D reconstruction methods commonly rely on coupled nonconvex optimization over camera poses, scene geometry, deformation, and appearance~\cite{lei25cvpr-mosca,wang25iccv-shapeofmotion}. Such formulations have achieved impressive results, but their strong coupling makes the optimization difficult: the objective is highly nonconvex and local optimization may therefore be sensitive to initialization and converge to undesired local minima. As illustrated in Fig.~\ref{fig:edg-local-min}, local optimization of the motion of an embedded deformation graph (EDG) can converge to an inferior solution whereas the certifiable solver presented in this work achieves an  accurate reconstruction. Similar failures of local optimization in static reconstruction were demonstrated in XM~\cite{han25rss-xm} (Fig.~10). This raises a natural question:

\begin{tcolorbox}[crgstatement]
\textit{Can we achieve accurate and efficient 4D reconstruction through globally certifiable geometric optimization?}
\end{tcolorbox}

\begin{figure*}[t]
\centering
\includegraphics[width=\textwidth]{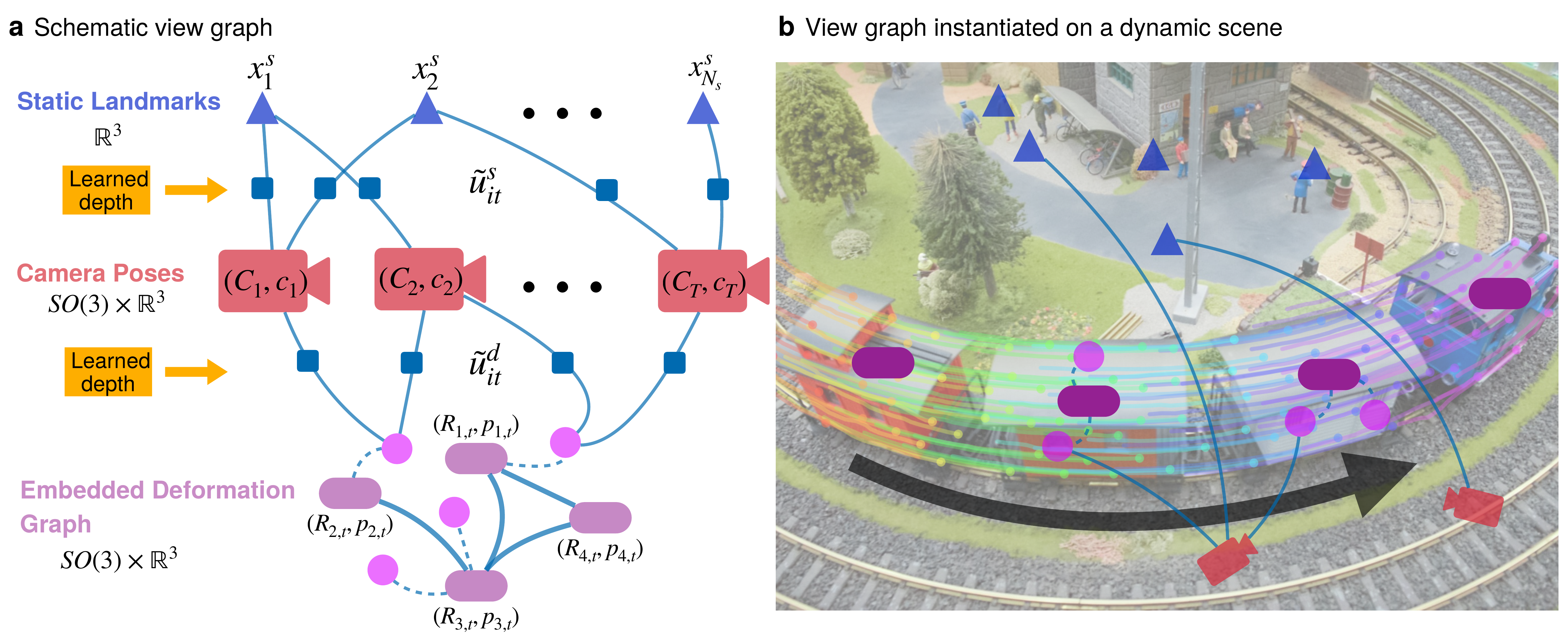}
\caption{\textbf{The \XMFD view graph.} (a) Schematic
view graph. Camera poses
$(C_t,\mathbf c_t)\in\mathrm{SO}(3)\times\mathbb R^{3}$ (red) observe
static landmarks $\mathbf x^{s}_i\in\mathbb R^{3}$ (blue triangles)
and dynamic points (pink
circles) through the depth-lifted measurements
$\tilde{\mathbf u}^{s}_{it}$ and $\tilde{\mathbf u}^{d}_{it}$ (blue
squares). Each dynamic keypoint is attached by
skinning edges (dashed) to its neighboring nodes
$(R_{j,t},\mathbf p_{j,t})$ of the embedded deformation graph (EDG)
(purple capsules), whose nodes are connected by EDG edges
(thick solid lines). (b) The view graph instantiated on a dynamic
scene: a train moves along a circular track (black arrow), colored
curves show the dynamic point tracks, and each part of the train is
an EDG node with a free $\mathrm{SE}(3)$ pose.}
\label{fig:view-graph}
\end{figure*}

In this work, we present \XMFD, a framework for accurate and efficient 4D reconstruction based on globally certifiable geometric optimization. Our approach separates camera motion estimation from non-rigid deformation, allowing both stages to be formulated as structured (nonconvex) quadratically constrained quadratic programs (QCQPs) and solved through (convex) semidefinite programming (SDP) relaxations. Fig.~\ref{fig:view-graph}(a) illustrates this decomposition as a view graph and Fig.~\ref{fig:view-graph}(b) instantiates it on a real scene: static and dynamic point tracks share the same cameras and predicted depth, but static tracks constrain time-invariant landmarks and camera poses, while dynamic tracks constrain the motion of an EDG. In the following, we overview the problem formulation and solution approach for both stages.

\textbf{I: Static stage.} As shown in Fig.~\ref{fig:view-graph}, let \(t\in\{1,\dots,T\}\) index video frames and \(i\in\{1,\dots,N_s\}\) index static points. The visibility set \(\mathcal E_s\) contains the tracked point-frame pairs, and \(\tilde{\mathbf u}^{s}_{it}\in\mathbb R^3\) denotes the corresponding 2D observation lifted to 3D using predicted monocular depth. Because monocular depth predictions are noisy and their scale drifts from frame to frame, we introduce a per-frame scale \(s_t>0\) that corrects this frame-dependent scale error in the predicted depth. Weighting each observation by an optional confidence \(\omega_{it}>0\), we jointly estimate camera poses, depth scales, and static landmarks via
\begin{equation}
\min_{\substack{
C_t\in\mathrm{SO}(3),\;\mathbf c_t \in \mathbb R^3,\\
s_t>0,\;\mathbf x_i^s \in \mathbb R^3
}}
\sum_{(i,t)\in\mathcal E_s}
\omega_{it}
\norm{
C_t(s_t\tilde{\mathbf u}_{it}^{s})
+\mathbf c_t-\mathbf x_i^s
}^2.
\tag{SBA}\label{eq:sba}
\end{equation}
The objective enforces that observations of the same static point, after correcting the predicted depth scale and transforming them into the world frame, agree on a common 3D landmark. Without a scale constraint, the objective admits a degenerate solution as all variables shrink toward zero. The original XM formulation~\cite{han25rss-xm} prevents this collapse by anchoring the first-frame scale to $s_1=1$, making the reconstruction overfit onto the first frame. We instead impose a uniform sphere constraint, $\sum_t s_t^2=T$, which fixes the global scale without privileging any frame.
We formulate this problem as a quadratically constrained quadratic program (QCQP), relax it to a semidefinite program (SDP), and solve it efficiently using Burer--Monteiro (BM) factorization together with a Riemannian staircase, following the certifiable optimization framework of XM and SE-Sync~\cite{han25rss-xm,rosen19ijrr-sesync}.

\textbf{II: Dynamic stage.} After fixing the recovered camera trajectory, we transform dynamic observations into the world frame and model non-rigid motion with an EDG. Each dynamic point $\mathbf x_i^d$ is warped by the rigid transformations $(R_{j,t},\mathbf p_{j,t})$ of its neighboring EDG nodes, where $\mathbf g_j$ denotes the rest position of node $j$ and $w_{ij}$ the skinning weight. Let $\mathbf y_{it}^{d}=C_t(s_t\tilde{\mathbf u}_{it}^{d})+\mathbf c_t$ denote the corresponding depth-lifted observation in the world frame. The tracking loss is
\begin{equation}
\begin{split}
\mathcal L_{\mathrm{track}}&=\sum_{(i,t)\in\mathcal E_d}\norm{\Phi_t(\mathbf x_i^d)-\mathbf y_{it}^{d}}^2,\\
\Phi_t(\mathbf x_i^d)&=\sum_{j\in\mathcal N(i)}w_{ij}\!\left[R_{j,t}(\mathbf x_i^d-\mathbf g_j)+\mathbf g_j+\mathbf p_{j,t}\right].
\end{split}
\label{eq:dynamic-reconstruction}
\end{equation}
We optimize the EDG node transformations with tracking loss, as-rigid-as-possible (ARAP) and temporal smoothness regularization. Unlike MoSca~\cite{lei25cvpr-mosca}, which models deformation through relative transformations between frames, we express all node transformations relative to a fixed rest frame. With fixed EDG structure and skinning weights, this choice makes the tracking, ARAP, and temporal smoothness objectives quadratic in the node transformations. Eliminating node translations then yields a rotation-only QCQP that admits the same SDP-based certifiable optimization framework as the static stage. The recovered geometry then initializes dynamic Gaussian splatting for rendering.

In summary, \XMFD makes the following contributions:
\begin{itemize}

    \item For static reconstruction, we build on the QCQP--SDP--BM framework of our conference version, XM~\cite{han25rss-xm}, retaining its efficient optimization and global optimality certificates. We introduce a uniform sphere constraint that resolves the scale ambiguity without anchoring any individual frame.

    \item For dynamic reconstruction, we design an EDG-based non-rigid reconstruction formulation that incorporates tracking, rigidity, and temporal smoothness into a structured QCQP. This formulation admits the same globally certifiable optimization framework.
    
    \item We build \XMFD, an accurate and efficient 4D reconstruction pipeline that combines certifiable geometric optimization with dynamic Gaussian splatting. Extensive experiments across seven benchmarks demonstrate leading camera accuracy, fast and scalable optimization, and the best average novel-view rendering performance on all three dynamic benchmarks (DyCheck, NVIDIA, and Syn4D).
\end{itemize}

\textbf{Outline.} The remainder of this paper is organized as follows.
Section~\ref{sec:related-work} reviews related work.
Section~\ref{sec:static} presents the static stage: the rotation-only QCQP and SDP relaxation, the
BM solver with its optimality certificate, and the
sphere constraint. Section~\ref{sec:dynamic} extends this machinery
to the EDG-based dynamic reconstruction and shows that it admits the
same certifiable solver. Section~\ref{sec:pipeline} describes the
complete pipeline.
Section~\ref{sec:experiments} evaluates camera accuracy, runtime, and
novel-view rendering on seven benchmarks. Proofs and implementation
details are given in the appendices.
Visualizations and code are available on the project page:
\begin{center}
    \url{https://computationalrobotics.seas.harvard.edu/XM4D/}.
\end{center}

%% file: sections/related-work.tex
\section{Related Work}
\label{sec:related-work}

\subsection{Dynamic Scene Reconstruction}

Existing dynamic scene reconstruction methods can be grouped by how they constrain the reconstruction. Rendering-based methods fit time-varying scene representations to images, while geometry-driven methods use measurements such as depth and point tracks to estimate structure and motion. Rendering-based methods represent dynamics using time-varying (volumetric) fields~\cite{lombardi19tog-neuralvolumes,li21cvpr-neuralsceneflow}, deformable canonical radiance fields and their higher-dimensional extensions~\cite{pumarola20arxiv-dnerf,park21iccv-nerfies,park21arxiv-hypernerf}, or moving Gaussians~\cite{luiten243dv-dynamicgaussian,wu24cvpr-4dgs,stearns24sigasia-dynamicmarbles}. Many of them incorporate geometric regularization, and some jointly estimate camera motion~\cite{liu23cvpr-rodynrf}. However, recovering object or part level $\mathrm{SE}(3)$ motion for downstream scene understanding generally requires additional motion estimation. Moreover, many methods rely on accurate camera poses, carefully designed initialization, and heavy per-scene optimization.

Geometry-based methods use geometric measurements and motion priors to constrain dynamic reconstruction. Earlier non-rigid structure-from-motion methods recover 3D shape from monocular point tracks using low-rank or subspace priors~\cite{garg13cvpr-dynreconstruct,zhu14cvpr-dynreconstructsubspace}, while RGB-D methods use depth observations with deformation graphs or volumetric models to track motion and reconstruct geometry~\cite{newcombe15cvpr-dynamicfusion,innmann16eccv-volumedeform,bozic21cvpr-neuraldeformgraph}. More recently, MoSca~\cite{lei25cvpr-mosca} lifts learned depth and point-track priors into a motion scaffold and anchors Gaussians to it for photometric reconstruction. Shape of Motion~\cite{wang25iccv-shapeofmotion} represents scene motion using shared $\mathrm{SE}(3)$ motion bases, with per-point weights combining the bases to recover persistent trajectories from monocular depth and long-range 2D tracks. \XMFD also uses explicit local $\mathrm{SE}(3)$ transformations as an EDG. Unlike them, \XMFD estimates camera poses and temporally coupled graph transformations from depth and tracks with a certifiable geometric solver, making local rigid motions explicit without involving Gaussians.

\subsection{Learning-Assisted Geometric Reconstruction}

Learned geometric priors have become important components of reconstruction pipelines. Monocular depth models~\cite{yang24cvpr-depthanything,piccinelli24cvpr-unidepth,hu24tpami-metric3d,bochkovskiy25iclr-depthpro,lin25arxiv-da3} predict dense scene depth, while point trackers~\cite{xiao24cvpr-spatracker,karaev25iccv-cotracker3,doersch24accv-bootstap,zholus25iccv-tapnext,lai26arxiv-cowtracker} establish correspondences across frames. Together, depth and correspondence estimates provide complementary constraints for geometric reconstruction: depth informs scene structure, while tracks connect observations over time.

Recent feed-forward methods predict richer 3D scene representations directly from images. DUSt3R~\cite{wang24cvpr-dust3r} predicts pointmaps for image pairs and aligns them in a shared coordinate frame, and MUSt3R~\cite{cabon25cvpr-must3r} extends it to many views, and VGGT~\cite{wang25cvpr-vggt} jointly predicts camera parameters, depth maps, pointmaps, and 3D point tracks. In \XMFD, learned depth and tracks are measurements for an explicit geometric optimization.

\subsection{Certifiable Geometric Optimization}

Certifiable geometric optimization solves nonconvex estimation problems, such as bundle adjustment, while verifying global optimality~\cite{lasserre06siopt-sparsity,yang23mp-stride,zhao26arxiv-advances}. Semidefinite relaxations provide such certificates for pose synchronization~\cite{rosen19ijrr-sesync,yu24ral-simsync}, outlier-robust registration~\cite{yang20tro-teaser}, and general polynomial formulations of robust estimation~\cite{yang22pami-certifiably}. 

Our prior conference work, \XM~\cite{han25rss-xm}, introduced depth-assisted scaled bundle adjustment with a scalable solver based on BM factorization~\cite{burer03mp-nonlinear} and Riemannian optimization~\cite{boumal15arxiv-bm}, which \XMFD extends to dynamic scenes.

%% file: sections/static.tex
\section{Static Reconstruction with XM}
\label{sec:static}
This section presents the static stage of \XMFD.
Section~\ref{sec:ba} reviews scaled bundle adjustment and
its QCQP reformulation
(Propositions~\ref{prop:rotation-only}--\ref{prop:qcqp});
Section~\ref{sec:xm-formulation} relaxes the QCQP to a convex SDP
(Proposition~\ref{prop:sdp}); Section~\ref{sec:bm} solves the SDP
with certified global optimality via the Riemannian
staircase method (Algorithm~\ref{alg:staircase}). Section~\ref{sec:sphere} highlights our main contribution compared to XM~\cite{han25rss-xm}: a
uniform sphere constraint on the depth scales that replaces the
first-frame anchor.

\subsection{Bundle Adjustment with Learned Depth}
\label{sec:ba}

The static observations originate from 2D correspondences, obtained
by point trackers~\cite{doersch24accv-bootstap,karaev25iccv-cotracker3,zholus25iccv-tapnext}
or by classical keypoint detection and
matching~\cite{lowe04ijcv-sift,schonberger16cvpr-colmap}, and $\mathbf{u}^{s}_{it}\in\mathbb{R}^{2}$ denotes the normalized
image coordinate of static point $i$ in frame $t$ obtained with the
camera intrinsics, which are either calibrated or estimated by a
foundation model~\cite{lin25arxiv-da3,wang25cvpr-vggt}. Each observation is lifted to a noisy 3D
measurement using its predicted depth $d_{it}>0$,
\begin{equation}
\tilde{\mathbf{u}}^{s}_{it}
=
d_{it}
\begin{bmatrix}
\mathbf{u}^{s}_{it}\\
1
\end{bmatrix}
\in\mathbb{R}^{3},
\label{eq:depth-lift}
\end{equation}
which constitutes the depth-lifted observation of
Eq.~\eqref{eq:sba}.

The translations $\{\mathbf{c}_t\}$ and landmarks
$\{\mathbf{x}^{s}_i\}$ in Eq.~\eqref{eq:sba} are
unconstrained variables and thus can be
eliminated analytically to get a reduced scaled-rotation-only problem. 

\begin{proposition}[Scaled-Rotation-Only Formulation, adapted from~\cite{han25rss-xm}]
\label{prop:rotation-only}
Problem~\eqref{eq:sba} is equivalent to
\begin{equation}
\begin{split}
\rho^\star_{\mathrm{SRO}}=
\min_{\substack{s_t>0,\;C_t\in\mathrm{SO}(3)\\ t=1,\dots,T}}
&\operatorname{tr}\!\left(Q U^\top U\right),\\
U=
&\begin{bmatrix}
s_1C_1 & \cdots & s_TC_T
\end{bmatrix},
\end{split}
\tag{SRO}\label{eq:sro}
\end{equation}
where $Q\in\mathbb{S}^{3T}$, $Q\succeq0$, is a data matrix built
from the lifted measurements through the Moore--Penrose
pseudo-inverse of the view-graph Laplacian. In
particular, the optimal values coincide:
$\rho^\star_{\mathrm{SBA}}=\rho^\star_{\mathrm{SRO}}$, where
$\rho^\star_{\mathrm{SBA}}$ denotes the optimal value
of~\eqref{eq:sba}. The closed form of $Q$,
the proof, and the recovery of the optimal translations and
landmarks are given in Appendix~\ref{app:rotation-only}.
\end{proposition}

The objective of~\eqref{eq:sro} vanishes as $U\to0$, a collapse of
the global scale. XM excludes it by anchoring the first frame; we
instead exclude it with the uniform sphere constraint
$\norm{U}_F^2=3T$ motivated in Section~\ref{sec:sphere}.

By defining the set of scaled orthogonal matrices through the
quadratic constraint
\begin{equation}
\begin{split}
\mathrm{sO}(3)
&=\left\{\bar R\in\mathbb{R}^{3\times3}\;\middle|\;\bar R^\top\bar R=\alpha I_3,\;\alpha\geq0\right\}\\
&=\left\{sR\;\middle|\;s\geq0,\;R\in\mathrm{O}(3)\right\},
\end{split}
\label{eq:so3-scaled}
\end{equation}
which contains the scaled rotation $\bar C_t=s_tC_t$ of each frame,
\eqref{eq:sro} reduces to the following QCQP.

\begin{proposition}[QCQP Formulation, adapted from~\cite{han25rss-xm}]
\label{prop:qcqp}
Consider the QCQP
\begin{equation}
\begin{split}
\rho^\star_{\mathrm{QCQP}}=
\min_{\substack{\bar C_t\in\mathrm{sO}(3)\\ t=1,\dots,T}}
&\operatorname{tr}\!\left(Q U^\top U\right),\\
U=
\begin{bmatrix}
\bar C_1 & \cdots & \bar C_T
\end{bmatrix},
&\qquad
\norm{U}_F^2=3T,
\end{split}
\tag{QCQP}\label{eq:qcqp}
\end{equation}
and let $U^\star=[\bar C_1^\star,\dots,\bar C_T^\star]$ be a global
optimizer. If $\det\bar C_t^\star>0$ for all $t$, then $U^\star$ is
a global minimizer of~\eqref{eq:sro} under the same
sphere constraint. The proof is given in
Appendix~\ref{app:qcqp}.
\end{proposition}

Otherwise, rotations in $\mathrm{SO}(3)$ are recovered by projecting
each block onto the scaled rotations through its singular value
decomposition (Appendix~\ref{app:rounding}).

\subsection{Convex SDP Relaxation}
\label{sec:xm-formulation}

Since local solvers for the QCQP may stall in suboptimal minima, we use a semidefinite relaxation~\cite[Chapter~3]{Yang24book-sdp}. The basic idea is
fairly simple: creating a matrix variable
\begin{equation}
X=U^\top U\in\mathbb{S}^{3T},
\label{eq:x-lift}
\end{equation}
which is quadratic in the original variable $U$, makes the
objective of~\eqref{eq:qcqp} linear in $X$, namely
$\operatorname{tr}(QU^\top U)=\langle Q,X\rangle$. The convex
relaxation then enforces the matrix variable $X$ through convex constraints. Partition $X$ into
$3\times3$ blocks $X_{tt'}$ with $t,t'\in\{1,\dots,T\}$. Since each
$\bar C_t=s_tC_t$,
\begin{equation}
X_{tt}
=
\bar C_t^\top\bar C_t
=
s_t^2 I_3
=
\alpha_t I_3,
\qquad
\alpha_t=s_t^2.
\label{eq:x-diag}
\end{equation}
All diagonal blocks play identical roles, and the sphere
constraint becomes the linear equality $\operatorname{tr}(X)=3T$. Ignoring the
rank-$3$ constraint on $X$ yields Shor's SDP relaxation.

\begin{proposition}[SDP Relaxation, adapted from~\cite{han25rss-xm}]
\label{prop:sdp}
The semidefinite program
\begin{equation}
\begin{aligned}
\rho^\star_{\mathrm{SDP}}=
\min_{X}
\quad&
\langle Q,X\rangle\\
\text{s.t.}\quad&
X\succeq0,\\
&
X_{tt}=\alpha_t I_3,
\qquad t=1,\ldots,T,\\
&
\operatorname{tr}(X)=3T,
\end{aligned}
\tag{SDP}\label{eq:sdp}
\end{equation}
where each scalar $\alpha_t$ is not a separate variable but the
common diagonal entry of the block $X_{tt}$, is a convex relaxation of~\eqref{eq:qcqp}, i.e.,
$\rho^\star_{\mathrm{SDP}}\leq\rho^\star_{\mathrm{QCQP}}$. If a
global minimizer $X^\star$ has $\operatorname{rank}(X^\star)=3$,
then it factorizes as $X^\star=(U^\star)^\top U^\star$ with
$U^\star=[\bar C_1^\star,\dots,\bar C_T^\star]\in\mathrm{sO}(3)^{T}$,
unique up to multiplication by $\mathrm{O}(3)$, and $U^\star$
is a global optimizer of~\eqref{eq:qcqp}. In this case, the relaxation is said to be
\emph{tight} or \emph{exact} (proof in Appendix~\ref{app:sdp}).
\end{proposition}

In practice the relaxation is empirically tight~\cite{han25rss-xm};
when a minimizer has rank above three, it is projected to rank three
through its three largest eigenpairs, a process known as \emph{rounding} (Appendix~\ref{app:rounding}).

\subsection{Burer--Monteiro Optimization}
\label{sec:bm}

Although~\eqref{eq:sdp} is convex, explicitly storing and
optimizing the $3T\times3T$ matrix $X$ becomes prohibitive for long
videos with a large $T$: interior-point solvers already struggle beyond a few
thousand frames~\cite{han25rss-xm}. When the relaxation is tight,
however, the optimal $X^\star$ has rank three
(Proposition~\ref{prop:sdp}), so its effective dimension is only
$3\times3T$. BM
factorization~\cite{burer03mp-nonlinear} exploits this low-rank
structure: for a chosen rank $r\geq3$, we substitute
\begin{equation}
X=U^\top U,
\qquad
U=
\begin{bmatrix}
U_1 & \cdots & U_T
\end{bmatrix}
\in\mathbb{R}^{r\times3T},
\label{eq:bm-factor}
\end{equation}
with blocks $U_t\in\mathbb{R}^{r\times3}$ into~\eqref{eq:sdp},
which yields the rank-$r$ BM problem
\begin{equation}
\begin{split}
\rho^\star_{\mathrm{BM},r}=
\min_{U\in\mathbb{R}^{r\times3T}}\ 
&\operatorname{tr}\!\left(QU^\top U\right)\\
\text{s.t.}\quad
&U_t^\top U_t=\alpha_tI_3,
\quad t=1,\dots,T,\\
&\textstyle\sum_{t=1}^{T}\alpha_t=T.
\end{split}
\tag{BM}\label{eq:bm-block}
\end{equation}

At first glance, \eqref{eq:bm-block} looks like a nonconvex
constrained problem, but its constraints define a smooth manifold.

\begin{proposition}[BM Riemannian Optimization, adapted from~\cite{han25rss-xm}]
\label{prop:bm-manifold}
Problem~\eqref{eq:bm-block} is equivalent to the
\emph{unconstrained} Riemannian optimization problem
\begin{equation}
\min_{(\mathbf{s},Y_1,\dots,Y_T)\in\mathcal{M}}
\operatorname{tr}\!\left(QU^\top U\right),
\qquad
U_t=s_tY_t,
\label{eq:bm-param}
\end{equation}
over the product manifold
\begin{equation}
\mathcal{M}
=
\mathbb{S}^{T-1}_{\sqrt{T}}
\times
\mathrm{St}(r,3)^{T},
\label{eq:bm-manifold}
\end{equation}
where
$\mathbb{S}^{T-1}_{\sqrt{T}}
=\{\mathbf{s}\in\mathbb{R}^{T}:\norm{\mathbf{s}}_2^2=T\}$
is the radius-$\sqrt{T}$ sphere,
$\mathrm{St}(r,3)=\{Y\in\mathbb{R}^{r\times3}:Y^\top Y=I_3\}$
is the Stiefel manifold, and $\alpha_t=s_t^2$. The proof is given
in Appendix~\ref{app:bm-manifold}.
\end{proposition}

We solve~\eqref{eq:bm-param} with the Riemannian trust-region
method with truncated conjugate gradients
(\textsc{Rtr-tCG})~\cite{boumal14jmlr-manopt}, using the C++/CUDA
implementation from XM~\cite{han25rss-xm}.

Three observations are in
order~\cite{burer03mp-nonlinear,Yang24book-sdp}. First,
$X=U^\top U\succeq0$ holds by construction, so the PSD constraint
disappears. Second, at $r=3$, problem~\eqref{eq:bm-block} is
exactly~\eqref{eq:qcqp} and hence nonconvex. Third, increasing the rank cannot increase the optimal objective value, since any feasible rank-$r$ factor remains feasible at rank $r{+}1$ by appending a zero row; hence
\begin{equation}
\rho^\star_{\mathrm{BM},3}\;\geq\;\rho^\star_{\mathrm{BM},4}\;\geq\;
\cdots\;\geq\;\rho^\star_{\mathrm{SDP}},
\label{eq:bm-chain}
\end{equation}
with equality $\rho^\star_{\mathrm{BM},r}=\rho^\star_{\mathrm{SDP}}$
as soon as $r\geq r^\star$, the minimum rank among optimal solutions
of~\eqref{eq:sdp}~\cite{burer03mp-nonlinear,Yang24book-sdp}. We
emphasize that~\eqref{eq:bm-chain} is a statement about
\emph{global} minima: each $\rho^\star_{\mathrm{BM},r}$ is the
optimal value of the nonconvex problem~\eqref{eq:bm-block}, which a
local Riemannian solver alone is not guaranteed to attain.
Attaining it---and thereby $\rho^\star_{\mathrm{SDP}}$---is
precisely the role of the Riemannian
staircase~\cite{boumal15arxiv-bm,rosen19ijrr-sesync}
developed below.

\textbf{Global optimality certification.} Write~\eqref{eq:sdp} in
standard primal form,
\begin{subequations}\label{eq:sdp-standard}
\begin{align}
\min_{X\in\mathbb{S}^{3T}}\quad
&\operatorname{tr}(QX)\\
\text{s.t.}\quad
&\langle A_i,X\rangle=b_i,\quad i=1,\dots,m,
\label{eq:sdp-standard-linear}\\
&X\succeq0.
\label{eq:sdp-standard-psd}
\end{align}
\end{subequations}
The associated dual standard SDP is
\begin{subequations}\label{eq:sdp-dual}
\begin{align}
\max_{y\in\mathbb{R}^{m}}\quad
&\sum_{i=1}^{m}b_iy_i\\
\text{s.t.}\quad
&Z(y):=Q-\sum_{i=1}^{m}y_iA_i\succeq0.
\label{eq:dual-cert}
\end{align}
\end{subequations}
Since Slater's condition holds ($X=I_{3T}$ is feasible
for~\eqref{eq:sdp}), strong duality holds
between~\eqref{eq:sdp-standard}
and~\eqref{eq:sdp-dual}~\cite{wolkowicz12book-handbook}.

\begin{theorem}[Dual Optimality Certificate, adapted from~\cite{han25rss-xm}]
\label{thm:certificate}
Let $U_r^\star$ be a locally optimal solution
of~\eqref{eq:bm-block} with $\alpha_t>0$ for all $t$. Then there
exists a unique multiplier $y_r\in\mathbb{R}^{m}$ satisfying the
first-order condition
\begin{equation}
Z(y_r)(U_r^\star)^\top=0.
\label{eq:first-order}
\end{equation}
If, furthermore, $Z(y_r)\succeq0$, then $U_r^\star$ is a global
optimizer of~\eqref{eq:bm-block}, and $(U_r^\star)^\top U_r^\star$,
$y_r$, $Z(y_r)$ are optimal for the primal--dual
pair~\eqref{eq:sdp-standard}--\eqref{eq:sdp-dual}.
The proof is given in Appendix~\ref{app:licq}.
\end{theorem}

Theorem~\ref{thm:certificate} gives a simple certification recipe:
solve the linear system~\eqref{eq:first-order} for $y_r$ (recall
$Z(y)$ is linear in $y$), form $Z(y_r)$, and check its positive
semidefiniteness. If the check fails, the next theorem provides an
escape route.

\begin{theorem}[Negative-Curvature Direction, adapted from~\cite{han25rss-xm}]
\label{thm:descent}
Let $U_r^\star$ be a local minimizer of~\eqref{eq:bm-block} with
multiplier $y_r$, and suppose $Z(y_r)\not\succeq0$. Let $v$ be an
eigenvector of $Z(y_r)$ associated with a negative eigenvalue.
Then, for the rank-$(r{+}1)$ problem,
\begin{equation}
D=
\begin{bmatrix}
0\\ v^\top
\end{bmatrix}
\quad\text{is a second-order descent direction at}\quad
\bar U=
\begin{bmatrix}
U_r^\star\\ 0
\end{bmatrix}:
\end{equation}
along the feasible retraction curve through $\bar U$ in the direction $D$, the objective $f$ satisfies $f'(0)=0$ and $f''(0)=2v^\top Z(y_r)v<0$.
The proof is given in Appendix~\ref{app:descent}.
\end{theorem}

Since the objective decreases to second order along $D$, the halving line search in
Algorithm~\ref{alg:staircase} terminates with a strictly lower
objective, and the optimization escapes the rank-$r$ local minimizer
by climbing one step of the rank ``staircase''.

\begin{theorem}[Global Convergence, adapted from~\cite{han25rss-xm}]
\label{thm:global}
Algorithm~\ref{alg:staircase} returns a globally optimal solution
of the SDP pair~\eqref{eq:sdp-standard}--\eqref{eq:sdp-dual},
regardless of the initialization $U_r^0$ (proof in Appendix~\ref{app:global}).
\end{theorem}

Theorem~\ref{thm:global} makes the approach initialization-free.
In practice, Algorithm~\ref{alg:staircase} terminates at $r=3$ or
$4$~\cite{han25rss-xm}, so the worst case $r=3T$ is never
approached and the scalability advantage of the low-rank
factorization is retained.

\begin{algorithm2e}[t]
\caption{Riemannian Staircase}
\label{alg:staircase}
\KwIn{data matrix $Q\in\mathbb{S}^{3T}$}
\KwOut{certified optimizer $U^\star$ of~\eqref{eq:bm-block}}
$r\gets3$;\quad $U_r^0\gets[\,I_3\ \cdots\ I_3\,]$\;
\While{\emph{true}}{
$U_r^\star\gets\textsc{Rtr-tCG}\big(Q,U_r^0\big)$\tcp*{local minimizer of~\eqref{eq:bm-block}}
solve~\eqref{eq:first-order} for $y_r$ and form $Z(y_r)$\;
\If{$Z(y_r)\succeq0$}{
\KwRet $U^\star=U_r^\star$\tcp*{certified global optimum}
}
$v\gets$ eigenvector of the most negative eigenvalue of $Z(y_r)$\;
$\alpha\gets1$;\quad $\bar U\gets\begin{bmatrix}U_r^\star\\ 0\end{bmatrix}$;\quad
$D\gets\begin{bmatrix}0\\ v^\top\end{bmatrix}$\;
$U_{r+1}\gets\operatorname{Retr}_{\bar U}(\alpha D)$\;
\While{$\operatorname{tr}\big(QU_{r+1}^\top U_{r+1}\big)\geq\operatorname{tr}\big(Q(U_r^\star)^\top U_r^\star\big)$}{
$\alpha\gets\alpha/2$;\quad
$U_{r+1}\gets\operatorname{Retr}_{\bar U}(\alpha D)$\;
}
$r\gets r+1$;\quad $U_r^0\gets U_{r+1}$\;
}
\end{algorithm2e}

\subsection{Anchoring via a Uniform Sphere Constraint}
\label{sec:sphere}

As discussed in Section~\ref{sec:ba}, the original XM formulation removes the scale and rotation ambiguities of~\eqref{eq:sba} by anchoring the first frame ($C_1=I_3$, $\mathbf{c}_1=\mathbf{0}$, $s_1=1$). This pins the reconstruction to the first depth prediction, so the solver overfits to it when the first-frame observations are noisy, and XM's scale-regularization adds a hyperparameter that must be tuned for each video. We instead fix the global scale with a \emph{constraint} rather than a penalty: the uniform sphere constraint
\begin{equation}
\sum_{t=1}^{T}s_t^2=T
\label{eq:sphere}
\end{equation}
excludes the collapse toward zero while treating every frame symmetrically. It is quadratic in the scales, so~\eqref{eq:qcqp} remains a QCQP, and after relaxation it is the trace constraint $\operatorname{tr}(X)=3T$ already in~\eqref{eq:sdp} and~\eqref{eq:bm-block}.

\textbf{Suboptimality gap.} Since the staircase runs with finite
tolerances, we report the relative gap
\begin{equation}
\eta=
\frac{\hat\rho-\rho_{\mathrm{d}}}
{1+|\hat\rho|+|\rho_{\mathrm{d}}|},
\qquad
\rho_{\mathrm{d}}=\mathbf{b}^\top y_r+3T\lambda_{\min}\!\left(Z(y_r)\right),
\label{eq:static-gap}
\end{equation}
where $\hat\rho=\operatorname{tr}(Q\hat U^\top\hat U)$ is the
objective of the solution $\hat U$ rounded to~\eqref{eq:sro} under the
sphere constraint (Appendix~\ref{app:rounding}), $\mathbf{b}=(b_1,\dots,b_m)$
and $Z(y)$ are defined in~\eqref{eq:sdp-standard}--\eqref{eq:dual-cert},
and $y_r$ solves~\eqref{eq:first-order}. By
Theorem~\ref{thm:subopt} (Appendix~\ref{app:subopt}),
$\hat\rho\geq\rho^\star_{\mathrm{SRO}}\geq\rho^\star_{\mathrm{SDP}}\geq\rho_{\mathrm{d}}$,
so $\eta\to0$ certifies that $\hat U$ is numerically globally optimal
for~\eqref{eq:sro}.
Table~\ref{tab:subopt_summary} reports $\eta$ per dataset: all
static solves satisfy $\eta<10^{-3}$, where for the partial sphere
constraint $\eta$ uses the trace-restricted bound of Appendix~\ref{app:subopt}.\looseness=-1

\begin{tcolorbox}
\begin{center}
\textbf{Summary}
\end{center}
The static stage lifts 2D point tracks to 3D with learned depth,
eliminates translations and landmarks to obtain~\eqref{eq:sro},
and relaxes it through~\eqref{eq:qcqp} to the convex
program~\eqref{eq:sdp}, solved by the Riemannian staircase with a
certificate of global optimality and a suboptimality
gap~\eqref{eq:static-gap}. Unlike XM, a sphere constraint fixes the global scale without anchoring any frame.
The recovered camera poses $(C_t,\mathbf{c}_t,s_t)$ define the
world frame in which dynamic tracks are lifted for the non-rigid
stage.
\end{tcolorbox}

%% file: sections/dynamic.tex
\section{Dynamic Reconstruction with XM}
\label{sec:dynamic}

We extend the globally certifiable optimization framework of Section~\ref{sec:static} to the dynamic part of the view graph. Section~\ref{sec:edg-model} introduces the EDG motion model and dynamic objective; Section~\ref{sec:dynamic-qcqp} analytically eliminates the unconstrained variables to obtain a rotation-only QCQP, as in the static case; and Section~\ref{sec:dynamic-solve} describes how the XM solver of Section~\ref{sec:bm} solves the resulting problem.

\subsection{EDG Model and the Dynamic Objective}
\label{sec:edg-model}

Dynamic motion is represented by an EDG~\cite{sumner07siggraph-edg}, $\mathcal{G}_{\mathrm{EDG}}=(\mathcal{V}_{\mathrm{EDG}},\mathcal{E}_{\mathrm{EDG}})$, consisting of $J$ sparse control nodes (Eq.~\eqref{eq:dynamic-reconstruction}). Each node $j$ has a rest position $\mathbf{g}_j\in\mathbb{R}^3$ and a rigid transformation $(R_{j,t},\mathbf{p}_{j,t})\in\mathrm{SO}(3)\times\mathbb{R}^3$ describing its motion at frame $t$. Dynamic points are attached to nearby nodes and deformed using linear blend skinning. In this section, the graph (its nodes, edges, rest positions, and skinning weights) is taken as given; Section~\ref{sec:pipe-edg} describes how it is built from the lifted dynamic tracks.

We first describe the EDG model for a single temporal window $\mathcal{W}$. Within this window, we choose a reference frame $t_0\in\mathcal{W}$, typically the frame with the most observed EDG nodes, and define the rest positions in that frame, which keeps the skinning offsets and ARAP constraints consistent. We anchor the node transformations at $t_0$ by
\begin{equation}
R_{j,t_0}=I_3,\qquad
\mathbf{p}_{j,t_0}=\mathbf{0},
\qquad j=1,\dots,J,
\label{eq:edg-rest}
\end{equation}
so that $\Phi_{t_0}$ is the identity. This is a local reference-frame convention for one solve; it does not require a single rest frame for the entire video (Section~\ref{sec:pipe-edg}).

Since the cameras and depth scales are fixed at their static-stage
estimates, the lifted observations $\mathbf{y}^{d}_{it}$ are
constants, and the dynamic problem estimates the node transformations alone,
\begin{equation}
\begin{split}
\min_{\substack{R_{j,t}\in\mathrm{SO}(3),\\ \mathbf{p}_{j,t}\in\mathbb{R}^{3}}}\quad
&\lambda_{\mathrm{track}}\,\mathcal{L}_{\mathrm{track}}
+\lambda_{\mathrm{ARAP}}\,\mathcal{L}_{\mathrm{ARAP}}
+\lambda_{\mathrm{temp}}\,\mathcal{L}_{\mathrm{temp}}.
\end{split}
\tag{DYN}\label{eq:dyn}
\end{equation}

Here $\mathcal{L}_{\mathrm{track}}$ is the tracking loss~\eqref{eq:dynamic-reconstruction}, which asks the warped rest positions to match the lifted observations;
the as-rigid-as-possible (ARAP) energy over the EDG edges,
\begin{equation}
\begin{split}
\mathcal{L}_{\mathrm{ARAP}}
&=\sum_{t=1}^{T}
\sum_{(j,k)\in\mathcal{E}_{\mathrm{EDG}}}
\norm{\mathbf{r}^{\mathrm{arap}}_{jk,t}}^{2},\\
\mathbf{r}^{\mathrm{arap}}_{jk,t}
&=R_{j,t}(\mathbf{g}_k-\mathbf{g}_j)
+\mathbf{g}_j+\mathbf{p}_{j,t}
-\mathbf{g}_k-\mathbf{p}_{k,t},
\end{split}
\label{eq:arap}
\end{equation}
asks the transformation of node $j$ to also explain the
motion of its neighbor $k$, and the temporal smoothness energy
\begin{equation}
\mathcal{L}_{\mathrm{temp}}
=\sum_{t=1}^{T-1}\sum_{j=1}^{J}
\norm{R_{j,t}-R_{j,t+1}}_F^{2}
+\norm{\mathbf{p}_{j,t}-\mathbf{p}_{j,t+1}}^{2}
\label{eq:temp}
\end{equation}
discourages large inter-frame deformation.

\begin{figure*}[t]
\centering
\includegraphics[width=\textwidth]{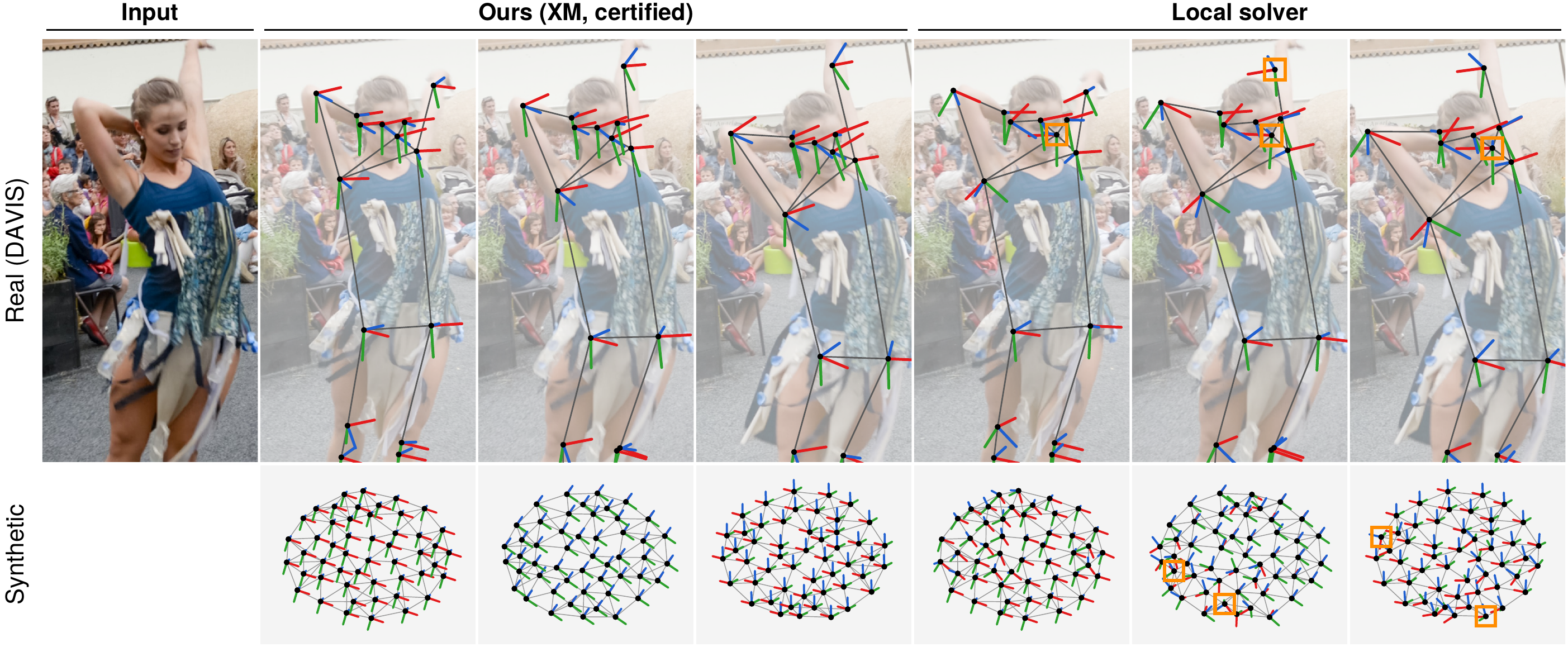}
\caption{\textbf{Certified versus local optimization.} Results on a
real DAVIS video~\cite{perazzi16cvpr-davis} (top) and a
synthetic deformation graph (bottom). Each panel shows the EDG nodes,
their estimated rotations (red, green, and blue axes), and the graph
edges at different frames. The local
solver converges to a local minimum with wrong node rotations
(orange boxes), whereas the certified XM solution recovers
consistent rotations.}
\label{fig:edg-local-min}
\end{figure*}

\subsection{Rotation-Only QCQP}
\label{sec:dynamic-qcqp}

Every residual in~\eqref{eq:dyn} is \emph{affine}: the tracking
residual is affine in $(R_{j,t},\mathbf{p}_{j,t})$ for fixed skinning weights and rest positions, the ARAP
residual is affine in $(R_{j,t},\mathbf{p}_{j,t},\mathbf{p}_{k,t})$
with the constant offset $\mathbf{g}_j-\mathbf{g}_k$, and the
smoothness residuals are linear. The objective is thus quadratic in the stacked variables, and~\eqref{eq:dyn} is
a QCQP whose only constraints are the orthogonality of the EDG
blocks.

As in the static case, the node translations $\mathbf{p}_{j,t}$ are
unconstrained Euclidean variables and can be eliminated
analytically, leaving a problem over the EDG rotations alone.

\begin{proposition}[Rotation-Only Formulation of the Dynamic Problem]
\label{prop:dyn-rotation-only}
Under the rest-frame convention~\eqref{eq:edg-rest},
Problem~\eqref{eq:dyn} is equivalent to
\begin{equation}
\begin{split}
\rho^\star_{\mathrm{DYN\text{-}RO}}=
\min_{\substack{R_{j,t}\in\mathrm{SO}(3)\\ t\neq t_0}}
&\operatorname{tr}\!\left(QU^\top U\right)
+2\,\langle P,U\rangle+c_0,\\
U=
&\begin{bmatrix}
R_{j,t}
\end{bmatrix}_{t\neq t_0},
\end{split}
\tag{DYN-RO}\label{eq:dyn-ro}
\end{equation}
where $Q\in\mathbb{S}^{3J(T-1)}$, $Q\succeq0$,
$P\in\mathbb{R}^{3\times3J(T-1)}$, and $c_0\in\mathbb{R}$ are built
from the lifted observations, the skinning weights, and the rest
positions through the Moore--Penrose pseudo-inverse of a
node-translation Laplacian. In particular, the
optimal values coincide: $\rho^\star_{\mathrm{DYN}}
=\rho^\star_{\mathrm{DYN\text{-}RO}}$, where
$\rho^\star_{\mathrm{DYN}}$ denotes the optimal value
of~\eqref{eq:dyn}. The closed forms of $Q$, $P$, and $c_0$,
the proof, and the recovery of the node translations are in Appendix~\ref{app:dyn-rotation-only}.
\end{proposition}

Two differences from the static
problem~\eqref{eq:sro} are worth noting. First, a \emph{linear}
term $\langle P,U\rangle$ appears: the ARAP and tracking residuals
carry the constant offsets $\mathbf{g}_j-\mathbf{g}_k$,
$\sum_j w_{ij}\mathbf{g}_j$, and $\mathbf{y}^{d}_{it}$, which
survive the elimination. Second, every block is a unit-scale
rotation, whose relaxed block constraint is $X_{jj}=I_3$ rather
than $\alpha I_3$, so no sphere normalization is needed.

\textbf{Homogenization.} The XM solver operates on a homogeneous
quadratic objective, so we absorb the linear term and the constant
of~\eqref{eq:dyn-ro} into a single matrix by introducing one auxiliary rotation
$V\in\mathrm{SO}(3)$ and defining
\begin{equation}
U_h=\begin{bmatrix}V & U\end{bmatrix},
\qquad
Q_h=
\begin{bmatrix}
\tfrac{c_0}{3}I_3 & P\\
P^\top & Q
\end{bmatrix}.
\label{eq:dyn-homog}
\end{equation}
The auxiliary block adds no physical degree of freedom; after solving,
the EDG rotations are taken relative to it.

\begin{proposition}[Homogeneous Formulation of the Dynamic Problem]
\label{prop:dyn-homog}
Problem~\eqref{eq:dyn-ro} is equivalent to
\begin{equation}
\rho^\star_{\mathrm{DYN\text{-}H}}=
\min_{\substack{V\in\mathrm{SO}(3),\;R_{j,t}\in\mathrm{SO}(3)\\
t\neq t_0}}
\operatorname{tr}\!\left(Q_hU_h^\top U_h\right),
\tag{DYN-H}\label{eq:dyn-h}
\end{equation}
in the following sense: the optimal values coincide,
$\rho^\star_{\mathrm{DYN\text{-}RO}}=\rho^\star_{\mathrm{DYN\text{-}H}}$;
if $(V^\star,U^\star)$ is optimal for~\eqref{eq:dyn-h}, then
$V^{\star\top}U^\star$ is optimal for~\eqref{eq:dyn-ro}; and if
$\widetilde U^\star$ is optimal for~\eqref{eq:dyn-ro}, then
$(I_3,\widetilde U^\star)$ is optimal for~\eqref{eq:dyn-h} (proof in Appendix~\ref{app:dyn-homog}).
\end{proposition}

\subsection{Solving with the XM Solver}
\label{sec:dynamic-solve}

As in Proposition~\ref{prop:qcqp}, we drop the determinant
constraints of~\eqref{eq:dyn-h}, replacing $\mathrm{SO}(3)$ by
$\mathrm{O}(3)$. The resulting
QCQP has orthogonal blocks only, so
Section~\ref{sec:bm} carries over with $n+1$ unit-scale blocks,
$n=J(T-1)$, and no sphere constraint.

\begin{proposition}[Certifiable Optimization of the Dynamic Problem]
\label{prop:dyn-sdp-bm}
The SDP
\begin{equation}
\begin{aligned}
\rho^\star_{\mathrm{DYN\text{-}SDP}}=
\min_{X\in\mathbb{S}^{3(n+1)}}\quad
&\langle Q_h,X\rangle\\
\text{s.t.}\quad
&X\succeq0,\\
&X_{ii}=I_3,\quad i=0,\dots,n,
\end{aligned}
\tag{DYN-SDP}\label{eq:dyn-sdp}
\end{equation}
is a convex relaxation of~\eqref{eq:dyn-h}. Its rank-$r$ BM factorization
$X=U_h^\top U_h$ is equivalent to the unconstrained Riemannian
problem
\begin{equation}
\min_{U_h\in\mathrm{St}(r,3)^{n+1}}
\operatorname{tr}\!\left(Q_hU_h^\top U_h\right).
\tag{DYN-BM}\label{eq:dyn-bm}
\end{equation}
Theorems~\ref{thm:certificate}--\ref{thm:global} still hold 
for~\eqref{eq:dyn-sdp}--\eqref{eq:dyn-bm}, so
Algorithm~\ref{alg:staircase} returns a certified global optimum
of~\eqref{eq:dyn-sdp}, and hence of~\eqref{eq:dyn-h} whenever the
relaxation is tight and the recovered relative rotation blocks lie in $\mathrm{SO}(3)$. The proof is given in
Appendix~\ref{app:dyn-sdp-bm}.\looseness=-1
\end{proposition}

Global optimality matters in practice. Fig.~\ref{fig:edg-local-min}
compares the certified solution with that of a local solver on the
same input: the local solver converges to a local minimum in which
some nodes are rotated incorrectly, whereas the certified solution
recovers consistent node rotations.

\textbf{Suboptimality gap.} The gap~\eqref{eq:static-gap} applies
verbatim to~\eqref{eq:dyn-sdp}, with $Q$ replaced by $Q_h$, $A_i$ and $b_i$ now encoding
$X_{ii}=I_3$, and $3T$ replaced by
$\operatorname{tr}(X)=3(n+1)$ (Appendix~\ref{app:subopt}). Every
segment solve on DyCheck, NVIDIA, and Syn4D is certified
(Table~\ref{tab:subopt_summary}). 

\begin{tcolorbox}
\begin{center}
\textbf{Summary}
\end{center}
The dynamic stage models non-rigid motion within each temporal segment with an EDG whose
transformations are expressed relative to a single rest frame, which
makes the tracking, ARAP, and temporal smoothness objective~\eqref{eq:dyn}
quadratic. Eliminating the node translations yields the rotation-only
problem~\eqref{eq:dyn-ro}, whose linear term is absorbed by one
auxiliary rotation into the homogeneous problem~\eqref{eq:dyn-h}. Its
SDP relaxation is solved with the same BM staircase as the
static stage, with a certificate of global optimality and a
suboptimality gap, whereas a local solver can stall in a wrong local
minimum.
\end{tcolorbox}

%% file: sections/pipeline.tex
\section{The \XMFD Pipeline}
\label{sec:pipeline}

This section describes the end-to-end system that recovers both
the static and the dynamic structure of the view graph in
Fig.~\ref{fig:view-graph}. We present it for a
monocular video, but the formulation only requires that each image
carry a time tag, and an image collection without time tags
reduces the pipeline to static 3D reconstruction. The stages are preprocessing and tracking
(Section~\ref{sec:pipe-pre}), the static stage
(Section~\ref{sec:pipe-static}), EDG construction,
segment solving and completion (Section~\ref{sec:pipe-edg}), and
the final photometric stage (Section~\ref{sec:pipe-photo}).

\subsection{Preprocessing: Depth, Intrinsics, Masks, and Tracks}
\label{sec:pipe-pre}

Every frame is processed by a monocular depth foundation
model,
with Depth Anything~3 (DA3)~\cite{lin25arxiv-da3} as the default,
which provides the 
metric depths $d_{it}$ used in the lift~\eqref{eq:depth-lift} and,
for uncalibrated videos, the camera intrinsics. Per-frame masks of the moving objects are obtained with
SAM~3~\cite{carion25arxiv-sam}, prompted either by a text
description or by a few hand-selected keypoints on the dynamic
object and then tracked across the video. Finally, both static and
dynamic 2D correspondences are produced by point
trackers,
with TAPNext~\cite{zholus25iccv-tapnext} as the default. The
trackers are operated in an \emph{online} fashion: as each frame
arrives, new query points are sampled only at locations farther
than a threshold from existing tracks. Unreliable observations
are discarded (Appendix~\ref{app:impl}). Optionally,
dynamic tracks can be assembled into long-range trajectories
for the EDG (Appendix~\ref{app:impl-long-tracks}).

\subsection{Static Stage: Certified Solve}
\label{sec:pipe-static}
The static stage solves~\eqref{eq:sba} with the solver of Section~\ref{sec:static}, augmented with two robustness steps. First,
we run XM$^2$~\cite{han25rss-xm}: the solver is applied twice, with an
outlier filter in between that keeps only the observations with
the smallest residuals under the first solution, so tracking
failures or inaccurate depth predictions do not corrupt the final estimate. Second, the result is
refined with Ceres~\cite{agarwal22software-ceres}, warm-started at the
certified XM$^2$ outputs and minimizing the classical 2D reprojection
error. Further implementation details of the
static stage are collected in Appendix~\ref{app:impl}.

\textbf{Partial sphere constraint.} In practice, some videos
contain a small fraction of outlier frames---strong blur or large
inter-frame motion, or depth predictions whose scale is far
off---and forcing such frames onto
the sphere lets them distort the global scale. We therefore use a
mixed-manifold variant of the solver: frames are ranked by the
trace of their diagonal block of $Q$, the top
$\lceil\rho T\rceil$ frames are constrained to the sphere of
Section~\ref{sec:sphere}, and the scales of the remaining frames
are optimized freely. Formally, with $\mathcal{S}$
the set of these frames, the sphere constraint becomes
$\sum_{t\in\mathcal{S}}s_t^2=|\mathcal{S}|$, i.e.,
$\sum_{t\in\mathcal{S}}\operatorname{tr}(X_{tt})=3|\mathcal{S}|$
in~\eqref{eq:sdp}, which is still linear, so the SDP relaxation and
the certificate of Theorem~\ref{thm:certificate} carry over.
The suboptimality gap~\eqref{eq:static-gap} is adapted
accordingly in Appendix~\ref{app:subopt}.
In the same spirit, we optionally normalize
the observation weights per frame: with
$w_t=\sum_{i}\omega_{it}$ the total weight of frame $t$, the
solver uses the raw weights $\omega_{it}$, equalizes every frame's
influence ($\omega_{it}/w_t$), or applies a softer normalization
($\omega_{it}/\sqrt{w_t}$). Table~\ref{tab:configurations} compares eight such configurations; configuration~(6) is our default.

\input{tables-main}

\subsection{EDG Construction, Segment Solve, and Completion}
\label{sec:pipe-edg}

Starting from the lifted dynamic tracks of
Section~\ref{sec:dynamic}, we first build a
time-independent graph connecting tracks that remain spatially close
over time. We then subsample this graph to obtain sparse EDG nodes
and inherit its connectivity to define the EDG edges. Because no
single frame observes all nodes in long videos, we solve the EDG
motion independently on short overlapping segments using the
certified solver of Section~\ref{sec:dynamic-solve}, and fuse
the segment solutions. Finally, missing EDG poses
are filled by a completion step.

\begin{figure*}[t]
\centering
\includegraphics[width=\textwidth]{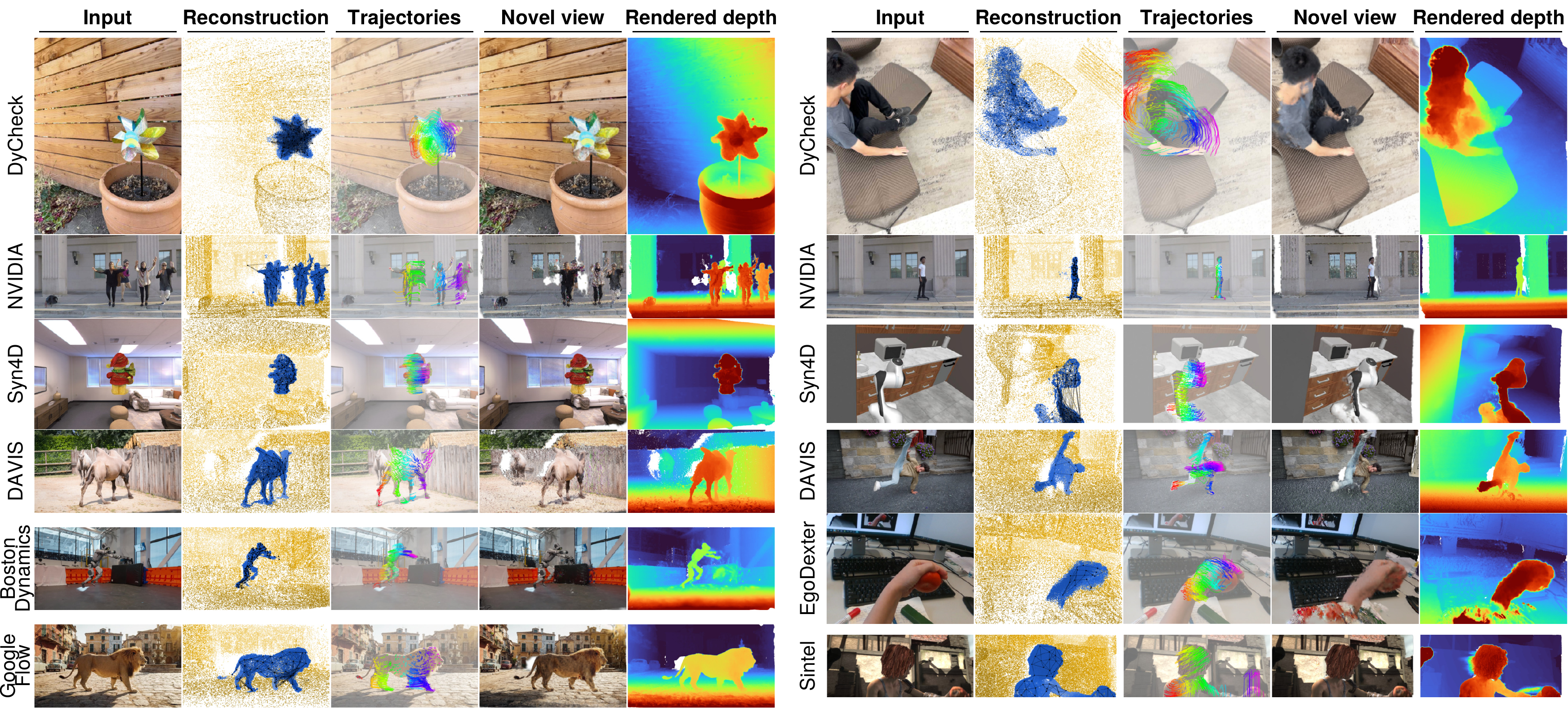}
\caption{\textbf{Qualitative results of \XMFD.} Sequences from DyCheck~\cite{gao22neurips-dycheck}, NVIDIA~\cite{yoon20cvpr-nvidiadataset}, Syn4D, DAVIS~\cite{perazzi16cvpr-davis}, Sintel~\cite{butler12eccv-sintel}, EgoDexter~\cite{mueller17iccv-egodexter}, a Boston Dynamics video~\cite{bd_atlas_airborne}, and a generated video from Google Flow. For each sequence, from left to right: an input frame, the reconstructed static and dynamic structure, the 3D trajectories of the dynamic
points over the video, a rendered novel view, and the rendered depth.}
\label{fig:qualitative}
\end{figure*}

\textbf{Track graph.} The track graph $\mathcal{G}_{\mathrm{tr}}$
records which tracks stay on the same part of the object. A
single-frame distance cannot decide this, since points on different
parts may pass close to each other in some frames. We therefore
measure the distance of two tracks over their whole common lifetime.
Let $\mathcal{T}_i$ be the frames in which track $i$ is reliable,
$\mathcal{T}_{ii'}=\mathcal{T}_i\cap\mathcal{T}_{i'}$ the frames in
which tracks $i$ and $i'$ are co-visible, and
$\boldsymbol{\pi}_{it}\in\mathbb{R}^2$ the pixel position of track
$i$ in frame $t$. We define the \emph{percentile distance} of two
tracks as the $\beta$-th percentile of their per-frame distance over
co-visible frames,
\begin{equation}
\begin{split}
d_{ii'}&=\operatorname{perc}_{\beta}
\big\{\norm{\mathbf{y}^{d}_{it}-\mathbf{y}^{d}_{i't}}\big\}_{t\in\mathcal{T}_{ii'}},\\
d^{\mathrm{2D}}_{ii'}&=\operatorname{perc}_{\beta}
\big\{\norm{\boldsymbol{\pi}_{it}-\boldsymbol{\pi}_{i't}}\big\}_{t\in\mathcal{T}_{ii'}},
\end{split}
\label{eq:percentile-dist}
\end{equation}
in the world frame and in the image, respectively. We write
$i\sim_{(\tau,\tau_{\mathrm{2D}})}i'$ if
\begin{equation}
\begin{gathered}
|\mathcal{T}_{ii'}|\geq\max\!\big(n_{\min},\,
\gamma\min(|\mathcal{T}_i|,|\mathcal{T}_{i'}|)\big),\\
d_{ii'}\leq\tau
\quad\text{or}\quad
d^{\mathrm{2D}}_{ii'}\leq\tau_{\mathrm{2D}},
\end{gathered}
\label{eq:track-accept}
\end{equation}
that is, the tracks overlap in time and stay close in 3D or 2D, and connect them in $\mathcal{G}_{\mathrm{tr}}$; the thresholds $\tau$ and $\tau_{\mathrm{2D}}$ are proportional to the median nearest-neighbor
distance of the tracks in 3D and in the image. Each track keeps at
most $D_{\max}$ edges with the smallest $d_{ii'}$.

\textbf{EDG nodes and edges.} The same criterion selects the EDG
nodes. Tracks are ranked by their length $|\mathcal{T}_i|$, and the
longest remaining track $j$ becomes a node and suppresses every
remaining track $i$ that stays on the same part of the object during
their common lifetime, i.e., with
$i\sim_{(\tau_{\mathrm{E}},\tau_{\mathrm{E,2D}})}j$
in~\eqref{eq:track-accept}, with thresholds and overlap fraction
that differ from those of the track graph, until every track is
either a node or suppressed by one. If this
yields more than $J_{\max}$ nodes, $J_{\max}$ of them are kept by
farthest-point sampling under the same percentile distance. Two nodes $j,j'$ are connected if they are within $k$ hops on the
track graph, which keeps independently moving parts apart. A node
left with fewer than three edges is connected to its nearest nodes
that are reachable on the track graph.

\textbf{Skinning.} Every other dynamic track $i$ is attached to
nodes $\mathcal{N}(i)$ chosen among the nodes within two hops of it
on the track graph; an isolated track with no node within two hops
falls back to its nearest node in percentile distance. The skinning
weights $w_{ij}$ are the normalized inverse distances from the track to its nodes at the rest frame. The graph and weights are fixed before solving, so the certificate covers the EDG motion given this graph.

\textbf{Segment solve and fusion.} For a long
sequence, we apply the single-window formulation of
Section~\ref{sec:dynamic} to overlapping windows $\{\mathcal{W}_k\}$.
Each window has its own reference frame $t_0^{(k)}$, rest positions,
and locally estimated transformations
$(R^{(k)}_{j,t},\mathbf{p}^{(k)}_{j,t})$, obtained by
solving~\eqref{eq:dyn} to certified global optimality with
Proposition~\ref{prop:dyn-sdp-bm}. We fuse the estimates over overlapping windows by averaging the estimated world-space node positions for each frame. We recover globally consistent rotations by synchronizing the relative rotations estimated within each window; see Appendix~\ref{app:impl-fusion}.

\textbf{Completion.} Node-frame pairs that remain empty after fusion
are filled by a local version of~\eqref{eq:dyn}: over all nodes and
frames, we minimize the same objective, with a robust 2D reprojection error as the tracking term, plus an initialization loss that keeps the solution close to the fused table, using a local solver initialized from it. This yields a complete
trajectory $(R_{j,t},\mathbf{p}_{j,t})$ for every node and frame.

\subsection{Photometric Stage}
\label{sec:pipe-photo}

The recovered geometry initializes a dynamic Gaussian-splatting
representation~\cite{kerbl23tog-3dgs}. We use
gsplat~\cite{ye25jmlr-gsplat} as the photometric training engine and add
an EDG wrapper on top of it, which warps every dynamic Gaussian to
frame $t$ by linear blend skinning with the EDG transformations.
During training, the rest positions of the dynamic
Gaussians are optimized together with the EDG transformations $(R_{j,t},\mathbf{p}_{j,t})$, initialized from the certified solve, under a photometric loss and depth supervision.

%% file: tables-main.tex

\begin{table}[t]
\centering
\caption{Mean camera accuracy ($\times10^{-3}$) of eight solver
configurations of Section~\ref{sec:pipe-static}, each determined by
the sphere fraction $\rho$ and the weight normalization:
(1)--(5)~raw weights at $\rho=0.1$, $0.3$, $0.5$, $0.9$, and $1$;
(6)~$\rho=0.9$ with the per-frame normalization $\omega_{it}/w_t$;
(7)~$\rho=0.9$ with the softer normalization
$\omega_{it}/\sqrt{w_t}$; and (8)~the original XM
solver~\cite{han25rss-xm} with raw weights. Configuration~6, our default, achieves the best or
second-best accuracy in eight of the twelve rows.}
\label{tab:configurations}
\small
\setlength{\tabcolsep}{3pt}
\begin{tabular}{llcccccccc}
\toprule
\multirow{2}{*}{Dataset} & \multirow{2}{*}{Metric} & \multicolumn{8}{c}{Configuration} \\
\cmidrule(lr){3-10}
 & & 1 & 2 & 3 & 4 & 5 & 6 & 7 & 8 \\
\midrule
\multirow{3}{*}{BAL}
& ATE-T & 2.243 & 2.194 & \underline{2.163} & 2.185 & 2.218 & \textbf{2.155} & 2.181 & 2.283 \\
& RPE-T & 4.003 & 3.942 & \textbf{3.934} & 3.936 & \underline{3.935} & 3.957 & 3.949 & 3.979 \\
& RPE-R & 1.674 & 1.661 & \textbf{1.650} & 1.667 & \underline{1.655} & 1.660 & 1.712 & 1.690 \\
\addlinespace[2pt]
\multirow{3}{*}{Sintel}
& ATE-T & 53 & 53 & 54 & \underline{51} & 52 & 56 & \textbf{49} & 74 \\
& RPE-T & 72 & \underline{71} & 75 & \textbf{69} & \underline{71} & 74 & \textbf{69} & 105 \\
& RPE-R & 76 & 71 & 70 & 66 & 67 & \textbf{52} & \underline{56} & 121 \\
\addlinespace[2pt]
\multirow{3}{*}{Sintel (Ceres)}
& ATE-T & 29 & 30 & 29 & \textbf{16} & \underline{18} & \underline{18} & 29 & 30 \\
& RPE-T & 35 & 37 & 37 & \textbf{25} & 28 & \underline{26} & 36 & 43 \\
& RPE-R & 62 & 63 & 61 & \underline{48} & 56 & \textbf{41} & 49 & 107 \\
\addlinespace[2pt]
\multirow{3}{*}{TUM (Ceres)}
& ATE-T & 9.45 & 8.80 & 10.48 & 8.70 & 9.21 & \textbf{5.72} & \underline{7.33} & 11.42 \\
& RPE-T & 12.85 & 12.22 & 14.51 & 12.42 & 12.99 & \textbf{8.02} & \underline{10.04} & 16.23 \\
& RPE-R & 6.63 & 6.18 & 7.69 & 6.28 & 6.64 & \underline{5.98} & \textbf{5.87} & 7.05 \\
\bottomrule
\end{tabular}
\end{table}

\begin{table}[t]
\centering
\caption{Camera accuracy averaged per dataset.
MoSca~\cite{lei25cvpr-mosca} crashes on IMC~\cite{Jin20ijcv-imc2023} (--), whose unordered photo
collections are not video data. VGGT~\cite{wang25cvpr-vggt} is run on subsampled frames on DyCheck~\cite{gao22neurips-dycheck} (Section~\ref{sec:exp-runtime}).}
\label{tab:accuracy-summary}
\small
\setlength{\tabcolsep}{3pt}
\begin{tabular}{llccccc}
\toprule
\multirow{2}{*}{Dataset} & \multirow{2}{*}{Metric} & \multicolumn{5}{c}{Method} \\
\cmidrule(lr){3-7}
 & & COLMAP & VGGT & MegaSAM & MoSca & Ours \\
\midrule
\multirow{3}{*}{IMC}
& ATE-T & \textbf{0.003} & 0.075 & 0.871 & -- & \underline{0.047} \\
& RPE-T & \textbf{0.005} & 0.121 & 1.221 & -- & \underline{0.074} \\
& RPE-R & \textbf{0.002} & \underline{0.033} & 0.703 & -- & 0.044 \\
\addlinespace[2pt]
\multirow{3}{*}{Replica}
& ATE-T & \underline{0.009} & 0.014 & 0.067 & 0.060 & \textbf{0.002} \\
& RPE-T & \underline{0.012} & 0.019 & 0.087 & 0.072 & \textbf{0.002} \\
& RPE-R & 0.015 & \underline{0.005} & 0.574 & 0.007 & \textbf{0.001} \\
\addlinespace[2pt]
\multirow{3}{*}{TUM}
& ATE-T & \textbf{0.006} & 0.008 & \underline{0.007} & 0.009 & \textbf{0.006} \\
& RPE-T & \underline{0.009} & 0.012 & \underline{0.009} & 0.013 & \textbf{0.008} \\
& RPE-R & \textbf{0.006} & \textbf{0.006} & \textbf{0.006} & \underline{0.007} & \textbf{0.006} \\
\addlinespace[2pt]
\multirow{3}{*}{Sintel}
& ATE-T & 0.359 & 0.066 & \underline{0.049} & 0.067 & \textbf{0.018} \\
& RPE-T & 0.462 & 0.095 & \underline{0.058} & 0.094 & \textbf{0.026} \\
& RPE-R & 0.081 & 0.057 & \textbf{0.019} & 0.056 & \underline{0.041} \\
\addlinespace[2pt]
\multirow{3}{*}{Syn4D}
& ATE-T & 0.035 & 0.009 & \textbf{0.001} & 0.022 & \underline{0.002} \\
& RPE-T & 0.060 & 0.011 & \textbf{0.001} & 0.031 & \underline{0.002} \\
& RPE-R & 0.022 & 0.012 & \textbf{0.001} & 0.057 & \underline{0.002} \\
\addlinespace[2pt]
\multirow{3}{*}{NVIDIA}
& ATE-T & 0.150 & 0.111 & \textbf{0.024} & 0.174 & \underline{0.038} \\
& RPE-T & 0.206 & 0.168 & \textbf{0.036} & 0.259 & \underline{0.056} \\
& RPE-R & 0.010 & 0.010 & \textbf{0.005} & 0.107 & \underline{0.005} \\
\addlinespace[2pt]
\multirow{3}{*}{DyCheck}
& ATE-T & \underline{0.006} & 0.019 & \textbf{0.005} & 0.048 & 0.008 \\
& RPE-T & \underline{0.008} & 0.026 & \textbf{0.007} & 0.065 & 0.011 \\
& RPE-R & \underline{0.009} & 0.035 & \textbf{0.007} & 0.024 & 0.012 \\
\bottomrule
\end{tabular}
\end{table}

\begin{table}[t]
\centering
\caption{Runtime of camera estimation. \emph{Top:} runtime per scene in
seconds, averaged per dataset and split into preprocessing (depth,
masks, tracking) and the solver.
\emph{Bottom:} wall-clock time in seconds on Replica~\cite{straub19arxiv-replica} room0 as the
number of input frames $N$ grows (contiguous prefix); a run is invalid
if it exceeds the GPU memory (\oom), runs longer than 1\,h
(\timeout), or crashes (\err). Best in bold, second best underlined.}
\label{tab:runtime-summary}
\small
\setlength{\tabcolsep}{3pt}
\begin{tabular}{llccccc}
\toprule
\multirow{2}{*}{Dataset} & \multirow{2}{*}{Stage} & \multicolumn{5}{c}{Method} \\
\cmidrule(lr){3-7}
 & & COLMAP & VGGT & MegaSAM & MoSca & Ours \\
\midrule
IMC & Total & 227.7 & \underline{38.7} & 136.8 & -- & \textbf{23.2} \\
\addlinespace[2pt]
\multirow{3}{*}{Replica}
& Total      & 125.4 & \textbf{54.4} & \underline{83.5} & 1054.4 & 115.1 \\
& Preprocess & -- & -- & \textbf{49.6} & 888.5 & \underline{100.5} \\
& Solver     & -- & -- & \underline{33.9} & 165.9 & \textbf{14.6} \\
\addlinespace[2pt]
\multirow{3}{*}{TUM}
& Total      & 1743.7 & \textbf{52.8} & \underline{79.5} & 403.8 & 83.7 \\
& Preprocess & -- & -- & \textbf{50.8} & 288.3 & \underline{73.7} \\
& Solver     & -- & -- & \underline{28.5} & 115.3 & \textbf{10.0} \\
\addlinespace[2pt]
\multirow{3}{*}{Sintel}
& Total      & 206.0 & \textbf{29.3} & \underline{48.4} & 410.2 & 62.0 \\
& Preprocess & -- & -- & \textbf{34.1} & 363.2 & \underline{47.2} \\
& Solver     & -- & -- & \textbf{14.3} & 47.0 & \underline{14.8} \\
\addlinespace[2pt]
\multirow{3}{*}{Syn4D}
& Total      & 605.0 & \textbf{55.0} & \underline{82.6} & 1729.3 & 110.3 \\
& Preprocess & -- & -- & \textbf{52.0} & 724.6 & \underline{88.4} \\
& Solver     & -- & -- & \underline{30.6} & 1004.6 & \textbf{21.9} \\
\addlinespace[2pt]
\multirow{3}{*}{NVIDIA}
& Total      & \underline{25} & \textbf{17} & 35 & 111 & 28 \\
& Preprocess & -- & -- & \underline{28} & 92 & \textbf{25} \\
& Solver     & -- & -- & \underline{7} & 18 & \textbf{3} \\
\addlinespace[2pt]
\multirow{3}{*}{DyCheck}
& Total      & 843 & \textbf{38} & 316 & 942 & \underline{96} \\
& Preprocess & 211 & -- & \underline{149} & 555 & \textbf{78} \\
& Solver     & 631 & -- & \underline{167} & 387 & \textbf{18} \\
\midrule
\multirow{6}{*}{\shortstack[l]{Replica\\room0}} & $N{=}100$ & \textbf{36.4} & \underline{37.6} & 89.6 & 468.5 & 73.7 \\
 & $N{=}200$ & 225.8 & \oom & \underline{149.0} & 892.6 & \textbf{131.3} \\
 & $N{=}500$ & 1451.9 & -- & \textbf{345.0} & 2140.1 & \underline{383.5} \\
 & $N{=}800$ & 3230.0 & -- & \textbf{568.6} & \err & \underline{862.3} \\
 & $N{=}1200$ & \timeout & -- & \err & -- & \textbf{1656.2} \\
 & $N{=}2000$ & -- & -- & -- & -- & \textbf{3322.4} \\
\bottomrule
\end{tabular}
\end{table}

\begin{table}[t]
\centering
\caption{Novel-view rendering averaged over the scenes of each
dataset; (w)/(w/o): with/without
ground-truth camera poses. \XMFD is the
best on every metric although it estimates its camera poses.}
\label{tab:rendering_summary}
\small
\setlength{\tabcolsep}{2pt}
\newcommand{\dyrowgap}{0.9pt}
\resizebox{0.8\columnwidth}{!}{%
\begin{tabular}{@{}c@{\hspace{8pt}}c@{}}
\toprule
\begin{tabular}[t]{@{}lccc}
\textbf{DyCheck} & mPSNR $\uparrow$ & mSSIM $\uparrow$ & mLPIPS $\downarrow$ \\
\midrule
T-NeRF (w) & 16.96 & 0.577 & 0.379 \\[\dyrowgap]
NSFF (w) & 15.46 & 0.551 & 0.396 \\[\dyrowgap]
Nerfies (w) & 16.45 & 0.570 & 0.339 \\[\dyrowgap]
HyperNeRF (w) & 16.81 & 0.569 & 0.332 \\[\dyrowgap]
DynPoint (w) & 16.89 & 0.573 & -- \\[\dyrowgap]
PGDVS (w) & 15.88 & 0.548 & 0.340 \\[\dyrowgap]
DyBluRF (w) & 17.37 & 0.591 & 0.373 \\[\dyrowgap]
CTNeRF (w) & 17.69 & 0.531 & -- \\[\dyrowgap]
Shape of Motion (w) & 17.32 & 0.598 & 0.296 \\[\dyrowgap]
Gaussian Marbles (w) & 16.72 & -- & 0.413 \\[\dyrowgap]
RoDynRF (w/o) & 17.10 & 0.534 & 0.517 \\[\dyrowgap]
MoSca (w/o) & \underline{18.91} & \underline{0.677} & \underline{0.284} \\[\dyrowgap]
Ours (w/o) & \textbf{19.51} & \textbf{0.725} & \textbf{0.163}
\end{tabular}
&
\begin{tabular}[t]{lcc@{}}
\textbf{NVIDIA} & PSNR $\uparrow$ & LPIPS $\downarrow$ \\
\midrule
NSFF (w) & 24.33 & 0.199 \\
HyperNeRF (w) & 17.60 & 0.367 \\
DynPoint (w) & 26.53 & 0.068 \\
CTNeRF (w) & 26.13 & 0.082 \\
RoDynRF (w) & 25.89 & \underline{0.067} \\
RoDynRF (w/o) & 25.38 & 0.079 \\
MoSca (w/o) & \underline{26.54} & 0.073 \\
Ours (w/o) & \textbf{27.01} & \textbf{0.063} \\
\midrule[\heavyrulewidth]
\textbf{Syn4D} & mPSNR $\uparrow$ & mLPIPS $\downarrow$ \\
\midrule
Gaussian Marbles (w) & 19.48 & 0.348 \\
Shape of Motion (w/o) & 16.72 & 0.418 \\
MoSca (w/o) & \underline{19.49} & \underline{0.235} \\
Ours (w/o) & \textbf{21.73} & \textbf{0.148}
\end{tabular}
\\
\bottomrule
\end{tabular}}
\end{table}

\begin{table}[t]
\centering
\caption{Per-scene runtime of the full dynamic reconstruction in minutes
(averages over the scenes of each dataset), split into preprocessing (depth, masks, static and dynamic
tracking) and optimization (camera estimation, the motion model, and
photometric training).
Best in bold.}
\label{tab:full-runtime}
\small
\setlength{\tabcolsep}{4pt}
\begin{tabular}{llccc}
\toprule
Dataset & Method & Preprocess & Optimization & Total \\
\midrule
\multirow{3}{*}{DyCheck} & Shape of Motion & 230.0 & 282.4 & 512.4 \\
 & MoSca & 9.2 & 31.3 & 40.5 \\
 & Ours & \textbf{8.8} & \textbf{25.8} & \textbf{34.6} \\
\addlinespace[2pt]
\multirow{3}{*}{NVIDIA} & Shape of Motion & \textbf{0.6} & 48.3 & 48.8 \\
 & MoSca & 1.5 & 32.1 & 33.6 \\
 & Ours & 1.5 & \textbf{20.4} & \textbf{21.9} \\
\addlinespace[2pt]
\multirow{3}{*}{Syn4D} & Shape of Motion & 20.0 & 50.9 & 71.0 \\
 & MoSca & 12.1 & 32.3 & 44.5 \\
 & Ours & \textbf{6.6} & \textbf{10.4} & \textbf{16.9} \\
\bottomrule
\end{tabular}
\end{table}

%% file: sections/experiments.tex
\section{Experiments}
\label{sec:experiments}

\begin{figure}[t]
\centering
\includegraphics[width=0.5\linewidth]{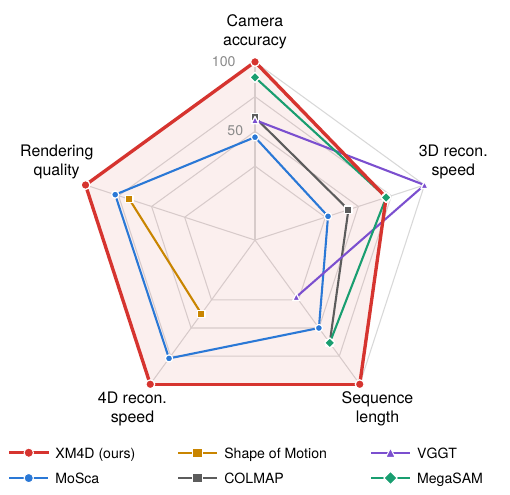}
\caption{\textbf{Performance overview.} Each axis is
the square-root-compressed ratio to the best method, averaged over the
datasets of Tables~\ref{tab:accuracy-summary}--\ref{tab:full-runtime}. Open polygons mark axes a method does not report. Farther out is
better. \XMFD performs best overall, leading on four of the five axes.}
\label{fig:radar}
\end{figure}

We compare \XMFD with Structure-from-Motion (SfM) and 4D reconstruction baselines on seven static and dynamic benchmarks in camera accuracy, runtime, and novel-view rendering, and extend the qualitative evaluation to additional videos (Fig.~\ref{fig:qualitative}); Fig.~\ref{fig:radar} summarizes the quantitative comparison.\looseness=-1

\subsection{Setup}
\label{sec:exp-setup}

\textbf{Datasets.} We use seven benchmarks. On four of
them---IMC~\cite{Jin20ijcv-imc2023},
Replica~\cite{straub19arxiv-replica,Zhu22cvpr-niceslam},
TUM~\cite{sturm12iros-tumdataset}, and
Sintel~\cite{butler12eccv-sintel}---we evaluate camera accuracy only. IMC and
Replica are static, while TUM and Sintel contain moving objects. On
the remaining three---Syn4D, NVIDIA~\cite{yoon20cvpr-nvidiadataset},
and the DyCheck iPhone dataset~\cite{gao22neurips-dycheck}---we
additionally run the EDG solve and the photometric reconstruction of
Section~\ref{sec:pipeline}. Syn4D is a synthetic benchmark that we
assemble from Replica~\cite{straub19arxiv-replica} and RoboCasa~\cite{nasiriany24rss-robocasa}. Each dataset is described in Appendix~\ref{app:datasets}. The solver-configuration study in Table~\ref{tab:configurations} additionally uses three problems from BAL~\cite{agarwal10eccv-bal}.

\textbf{Baselines.} For camera estimation, we compare against the classical SfM pipeline COLMAP~\cite{schonberger16cvpr-colmap}, the feed-forward geometry model VGGT~\cite{wang25cvpr-vggt}, and MegaSAM~\cite{li25cvpr-megasam}, which adapts DROID-SLAM~\cite{teed21neurips-droid} for camera pose and depth estimation in casual dynamic videos. For dynamic reconstruction, we compare against the geometry-based methods MoSca~\cite{lei25cvpr-mosca} and Shape of Motion~\cite{wang25iccv-shapeofmotion}. For novel-view rendering, we additionally compare against rendering-based methods that use time-varying fields (T-NeRF~\cite{gao22neurips-dycheck}, NSFF~\cite{li21cvpr-neuralsceneflow}, CTNeRF~\cite{miao24pr-ctnerf}, DyBluRF~\cite{bui23arxiv-dyblurf}, and RoDynRF~\cite{liu23cvpr-rodynrf}), deformable canonical radiance fields (Nerfies~\cite{park21iccv-nerfies}, HyperNeRF~\cite{park21arxiv-hypernerf}), or moving Gaussians (Gaussian Marbles~\cite{stearns24sigasia-dynamicmarbles}), and against DynPoint~\cite{zhou23neurips-dynpoint} and PGDVS~\cite{zhao24iclr-pgdvs}, which render from depth-warped points. 

\textbf{Hardware.} All experiments run on a workstation with an AMD
Ryzen Threadripper PRO 7985WX CPU (64 cores, 128 threads), 256\,GB
of RAM, and two 48\,GB NVIDIA RTX~6000 Ada GPUs, under Ubuntu 22.04 with CUDA 12.8. Every method, including ours, runs on a single GPU.

\textbf{Metrics.} We report the absolute trajectory error in
translation (ATE-T) and the relative pose error in translation and
rotation (RPE-T, RPE-R). The estimated camera centers
$\hat{\mathbf{c}}_t$ are first aligned to the ground-truth centers
$\mathbf{c}_t$ by a similarity transformation $(\sigma,R,\mathbf{t})$
estimated robustly with TEASER++~\cite{yang20tro-teaser}. The translation
errors are normalized by $d+1$, where
$d=\frac{1}{T}\sum_t\norm{\mathbf{c}_t-\bar{\mathbf{c}}}$ is the
mean distance of the ground-truth centers from their centroid:
\begin{equation*}
\begin{split}
\text{ATE-T}&=\frac{1}{d+1}\,\frac{1}{T}\sum_{t}
\norm{\sigma R\hat{\mathbf{c}}_t+\mathbf{t}-\mathbf{c}_t},\\
\text{RPE-T}&=\frac{1}{d+1}\,\frac{1}{|\mathcal{P}|}\sum_{(t,t')\in\mathcal{P}}
\norm{(\mathbf{c}_{t}-\mathbf{c}_{t'})-\sigma R(\hat{\mathbf{c}}_{t}-\hat{\mathbf{c}}_{t'})},
\end{split}
\end{equation*}
where $\mathcal{P}$ contains all frame pairs, and RPE-R is the mean
angle, in radians, of the relative rotation error over the same
pairs. The constant~$1$, in ground-truth units, prevents division by a
near-zero $d$ when the camera barely moves. For novel-view synthesis on DyCheck, NVIDIA, and Syn4D we
report mPSNR, mSSIM, and mLPIPS or their alternates. Following
MoSca~\cite{lei25cvpr-mosca}, we refine each test camera photometrically against its
test image with the scene frozen.

\subsection{Camera Accuracy}
\label{sec:exp-accuracy}

\begin{figure*}[p]
\centering
\newcommand{\figfivewidth}{\textwidth}\newcommand{\figsixwidth}{0.82\textwidth}%
\includegraphics[width=\figfivewidth]{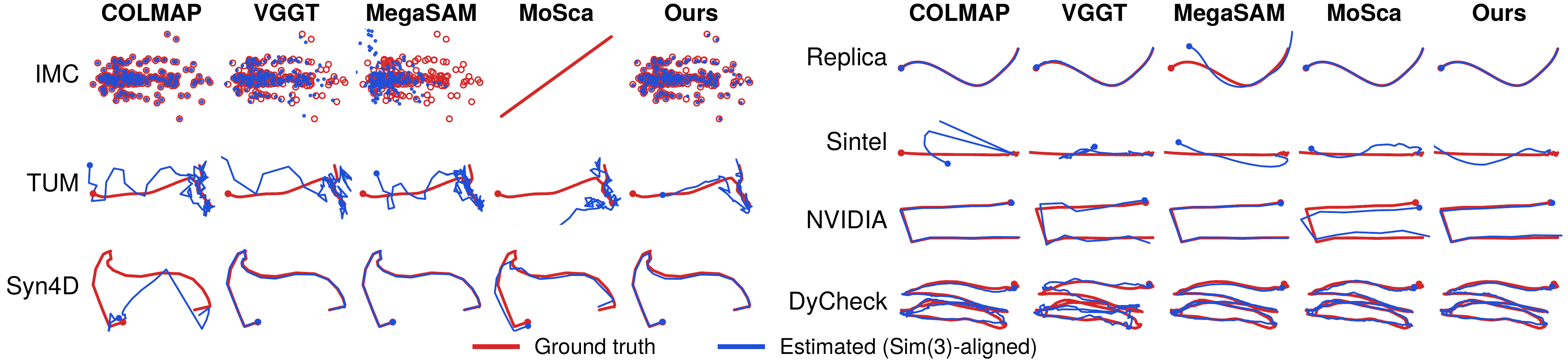}
\caption{\textbf{Estimated camera trajectories.} Estimated camera poses (blue) after Sim(3) alignment to the ground truth (red), shown for one scene per benchmark. IMC~\cite{Jin20ijcv-imc2023} consists of unordered images, whereas all other benchmarks contain temporally ordered sequences.}
\label{fig:trajectories}

\vspace{4pt}
\captionsetup[subfloat]{
  position=bottom,format=plain,labelformat=simple,labelsep=space,
  font=small,labelfont={bf,sf,color=black},textfont={footnotesize,sf,color=black},
  singlelinecheck=false,justification=centering,
  farskip=0pt,captionskip=1pt
}
\subfloat[DyCheck]{\includegraphics[width=\figsixwidth]{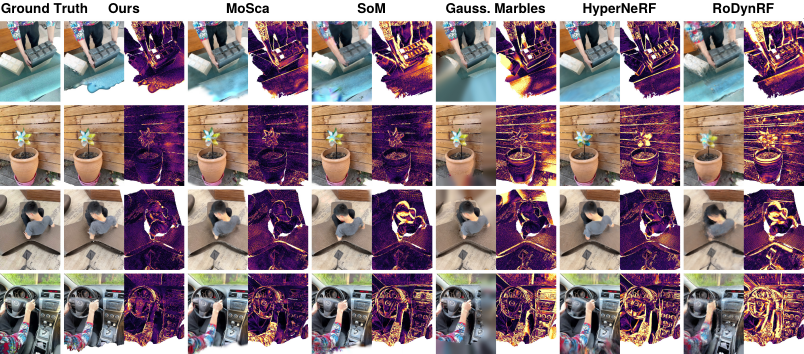}\label{fig:render-dycheck}}\\
\subfloat[NVIDIA]{\includegraphics[width=\figsixwidth]{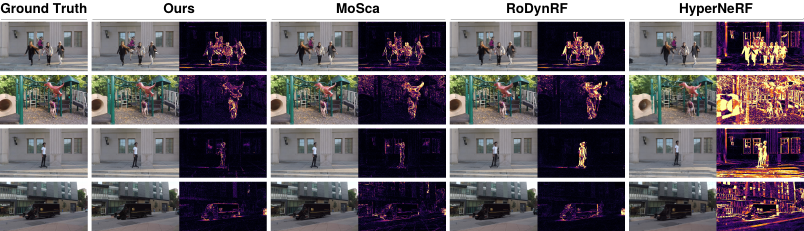}\label{fig:render-nvidia}}\\
\subfloat[Syn4D]{\includegraphics[width=\figsixwidth]{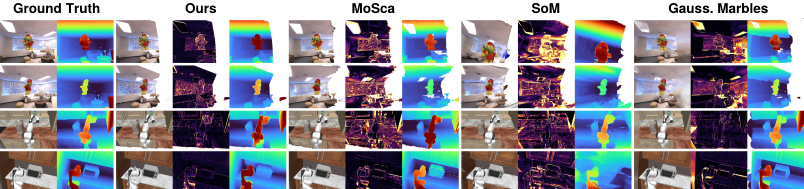}\label{fig:render-syn4d}}
\caption{\textbf{Novel-view rendering on DyCheck~\cite{gao22neurips-dycheck}, NVIDIA~\cite{yoon20cvpr-nvidiadataset}, and Syn4D.}
For each method, the rendered test view is shown with its error map
with respect to the ground truth (brighter is larger error).
(a)~DyCheck: from top to bottom, block, paper-windmill, spin, and
wheel, with error maps displayed only inside the co-visibility mask.
The other methods score lower mainly because inaccurate 3D geometry
misaligns their renderings with the test view, and because they blur
the dynamic parts. (b)~NVIDIA: from top to bottom, Jumping,
Playground, Skating, and Truck, with error maps over the full frame.
(c)~Syn4D: from top to bottom, two Bear~\cite{artec_christmas_bear} and two RoboCasa~\cite{nasiriany24rss-robocasa} sequences,
with the rendered depth also shown.}
\label{fig:render-compare}
\end{figure*}

Table~\ref{tab:accuracy-summary} summarizes camera accuracy across
all seven benchmarks. \XMFD is the most accurate method on Replica and TUM, and in translation on Sintel; on Syn4D, NVIDIA, and DyCheck it
is close to the best method, MegaSAM; and on the unordered image collections of IMC, \XMFD is outperformed only by the dedicated SfM pipeline COLMAP. Fig.~\ref{fig:trajectories} shows the trajectories on one scene per benchmark.

COLMAP is the most accurate on IMC, the unordered setting it is designed for, and remains
competitive on Replica and TUM, but degrades sharply on
Sintel, Syn4D, and NVIDIA, where its ATE-T is $4$--$20\times$ that of \XMFD. The
feed-forward VGGT is roughly three times less accurate in translation. MegaSAM, designed for dynamic videos,
is the most accurate on Syn4D, NVIDIA, and DyCheck, with \XMFD close
behind, but is less accurate on Replica and TUM and in translation on Sintel and fails on
IMC, whose images lack the temporal continuity its optical-flow front
end relies on. MoSca is $3$--$10\times$ less accurate than \XMFD in
ATE-T on the dynamic benchmarks and also fails on IMC; we attribute the
gap to the \XM solver, which recovers the cameras by a certifiably
optimal SDP on static tracks alone, so errors in the deformation model
cannot propagate into the trajectory. Overall, \XMFD ranks first on Replica and TUM and in translation on Sintel, second on Syn4D, NVIDIA, and in translation on IMC.

\subsection{Runtime}
\label{sec:exp-runtime}

\textbf{Static stage.} Table~\ref{tab:runtime-summary} reports the
runtime of camera estimation and,
where available, its split into preprocessing (depth, masks, and tracking) and the solver.
End to end, \XMFD is faster than MoSca on every benchmark and up to
$9\times$ faster than COLMAP on the long sequences, while the feed-forward VGGT is
the fastest overall, as expected for a single network pass. Against
MegaSAM the totals are mixed: \XMFD is faster on IMC, NVIDIA, and
DyCheck, and slower on Replica, TUM, Sintel, and Syn4D.

The preprocessing relies on foundation
models and will speed up as they improve, so we focus on the
solver, where \XMFD is the fastest on every benchmark but Sintel.

\textbf{Dynamic stage.} Table~\ref{tab:full-runtime} reports the
runtime of the full dynamic reconstruction against MoSca and Shape of
Motion. \XMFD is the fastest end to end and in optimization on all three benchmarks. Shape of Motion is
the slowest, mainly because its dense point tracking takes about four
hours per scene on DyCheck.\looseness=-1

\textbf{Sequence length.} Table~\ref{tab:runtime-summary} (bottom) reports the runtime on Replica room0 versus the number of frames $N$.
VGGT runs out of GPU memory at
$N=200$, MoSca fails beyond $N=500$, and beyond $N=800$ COLMAP
exceeds one hour and MegaSAM its $1024$-frame buffer, whereas \XMFD
completes every length, reaching $N=2000$ in $55$ minutes. VGGT's
limit is also why it runs on subsampled frames on DyCheck, whose
sequences exceed $100$ frames.

\subsection{Novel-View Rendering}
\label{sec:exp-rendering}

\begin{figure}[t]
\centering
\includegraphics[width=0.72\linewidth]{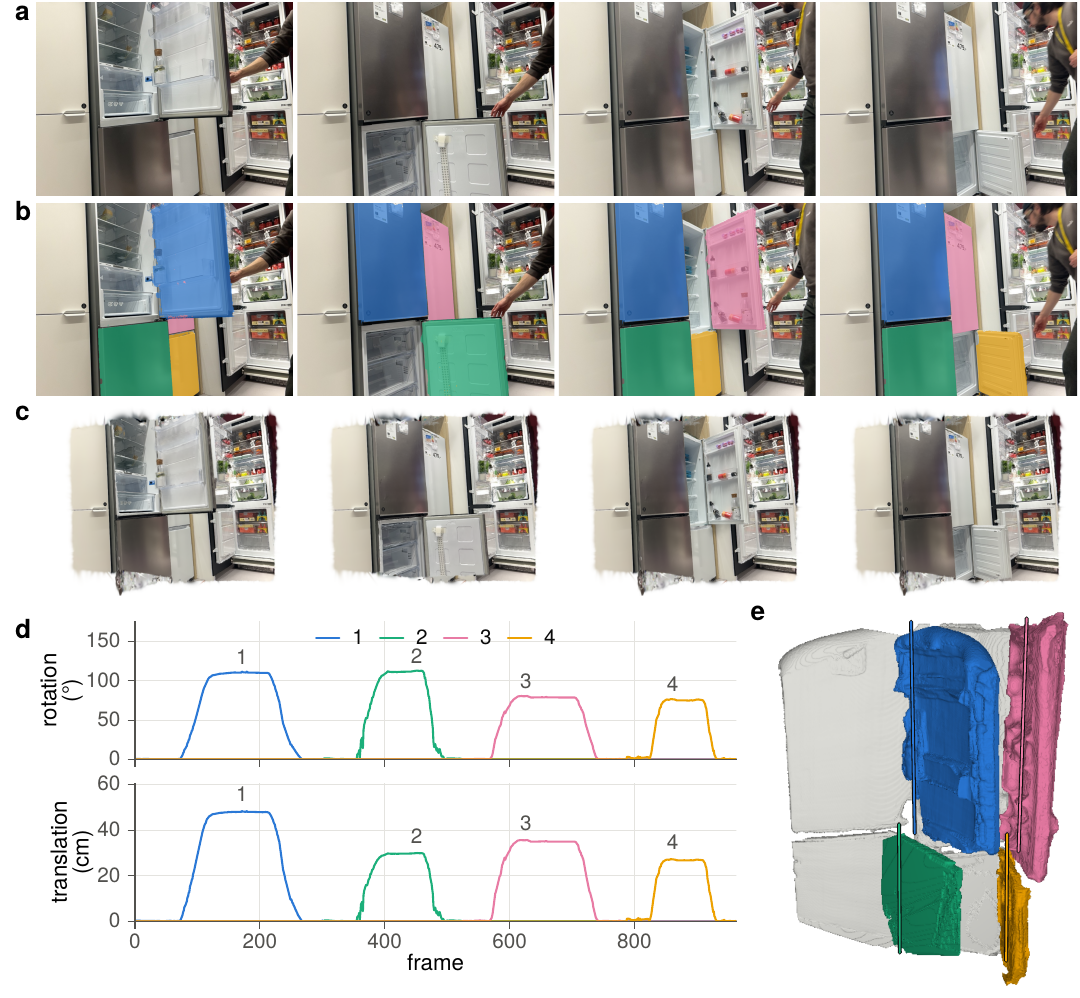}
\caption{\textbf{Articulated object reconstruction with \XMFD.} A
person opens the four doors of a refrigerator in turn in a casual
monocular video. (a)~Input frames. (b)~The four moving doors,
color-coded. (c)~Rendered views of the reconstruction at the same
time instants. (d)~Rotation angle and translation of each door over
the video. (e)~Reconstructed geometry with the four doors and their
estimated hinge axes.}
\label{fig:fridge}
\end{figure}

Table~\ref{tab:rendering_summary} and Fig.~\ref{fig:render-compare}
evaluate novel-view synthesis on DyCheck, NVIDIA, and Syn4D. \XMFD
attains the best average score on every metric of all three datasets. This holds
although most baselines are given the ground-truth camera poses (all
except MoSca and RoDynRF, and Shape of Motion on Syn4D), and Gaussian Marbles on Syn4D additionally the
ground-truth depth, whereas \XMFD estimates all cameras. The margins
are largest on Syn4D, where \XMFD improves PSNR by more than $2$\,dB
over the best baseline, and on DyCheck, where it has the best mSSIM and
mLPIPS on all seven scenes and the best mPSNR on three. As
Fig.~\ref{fig:render-compare} shows, the baselines lose mainly because
inaccurate geometry misaligns their renderings with the test view and
blurs the dynamic parts. Since the recovered dynamic motion is hard to
compare directly across methods, we further evaluate the geometry on
Syn4D, whose moving objects have ground-truth geometry, through the
Chamfer distance of the dynamic object and the error of the rendered
depth (Table~\ref{tab:syn4d-geometry}); \XMFD is the most accurate on
every scene.

\subsection{Qualitative Results}
\label{sec:exp-qualitative}

Fig.~\ref{fig:qualitative} shows \XMFD on dynamic videos beyond the benchmarks: in-the-wild videos from DAVIS~\cite{perazzi16cvpr-davis}, Sintel~\cite{butler12eccv-sintel}, egocentric hand--object interaction videos from EgoDexter~\cite{mueller17iccv-egodexter}, a Boston Dynamics humanoid robot performing a flip~\cite{bd_atlas_airborne}, and a generated video of a walking lion from Google Flow. These videos span real, synthetic, and generated footage, third-person and egocentric viewpoints, and the motion of animals, people, hands, and a robot; \XMFD handles all of them, demonstrating the robustness of the pipeline.

For videos of human body and hands, we build the EDG from keypoints instead of subsampling the track graph. For hands, WiLoR~\cite{potamias25cvpr-wilor} predicts a MANO~\cite{romero22arxiv-mano} hand in every frame; we keep only its 21 3D joints and connect them by the kinematic skeleton of the hand. For bodies, RTMPose~\cite{jiang23arxiv-rtmpose} detects the 17 COCO~\cite{lin14eccv-coco} keypoints in 2D within the bounding box of the person's mask.

Because the EDG represents motion with explicit $\mathrm{SE}(3)$ transformations, \XMFD recovers the part-level rigid motion of objects without an extra motion-estimation step. Such part-level articulation is what simulation-ready digital twins of articulated objects require, especially in robotics~\cite{jiang22cvpr-ditto,zhu23arxiv-groot,le25iclr-articulate,qi26icra-compose,hollis26arxiv-sceneagent}. Fig.~\ref{fig:fridge} shows a refrigerator whose four doors are opened in turn: the reconstruction yields the opening angle and displacement of every door over time and its hinge axis.

\begin{tcolorbox}
\begin{center}
\textbf{Summary}
\end{center}
\XMFD gives the best overall results for three reasons. First, its certified cameras make it the most accurate on Replica and TUM and in translation on Sintel, and close to the best elsewhere. Second, its solver is the fastest on nearly every benchmark, making \XMFD the fastest 4D reconstruction end to end, and scales to $2000$ frames. Third, its cameras and EDG motion give the photometric stage an accurate scaffold, yielding the best novel views on all three dynamic benchmarks, even against methods with ground-truth poses.
\end{tcolorbox}

%% file: sections/conclusion.tex
\section{Conclusions}
\label{sec:conclusions}

We presented \XMFD, a 4D reconstruction framework that recovers camera motion and non-rigid deformation through certifiably globally optimal geometric optimization. Separating the two stages turns both into QCQPs over (scaled) rotations: the static stage estimates camera poses, depth scales, and static landmarks, and the dynamic stage estimates the motion of an embedded deformation graph. Both QCQPs are relaxed to SDPs and solved by a GPU solver that returns a certificate of global optimality. Across seven benchmarks, \XMFD estimates accurate cameras on both dynamic videos and unordered photo collections, scales to thousands of frames, and initializes dynamic Gaussian splatting for state-of-the-art novel-view rendering.

\textbf{Limitations.} The certificate covers the optimization, not its inputs: \XMFD relies on the predicted monocular depth and point tracks, and errors that survive the filters of Appendix~\ref{app:impl-filter} bias the certified solution. The solver stores the data matrices densely, so its memory grows quadratically with the number of frames; exploiting their sparsity~\cite{papalia26icra-sparseavariable} is a natural next step. Finally, the EDG structure and skinning weights are fixed before the solve; estimating the graph jointly with its motion is future work.

%% file: sections/acknowledgements.tex
\section*{Acknowledgements}

This work was partially funded by Office of Naval
Research grant N00014-25-1-2322.

%% file: sections/app-tables.tex
\section{Additional Tables}
\label{app:tables}

Table~\ref{tab:subopt_summary} reports the suboptimality gaps per dataset.
Table~\ref{tab:syn4d-geometry} compares the reconstructed geometry on
Syn4D with MoSca~\cite{lei25cvpr-mosca}, Shape of
Motion~\cite{wang25iccv-shapeofmotion}, and Depth Anything~3~\cite{lin25arxiv-da3}.
For the Chamfer distance, the dynamic object at frame~$t$
is represented by the centers of the dynamic Gaussians at~$t$ (for
\XMFD, warped by the EDG); $1000$ points
are drawn uniformly at random from them and from the ground-truth
object point cloud, and the distance is the sum of the two mean
squared nearest-neighbor distances, averaged over the $100$ frames.

\input{tables-appendix}

%% file: tables-appendix.tex
\begin{table}[!h]
\centering
\caption{Median relative suboptimality $\eta$ of~\eqref{eq:static-gap} over every scene of each dataset.
$n_{\mathrm{ok}}$ counts solves with $\eta<10^{-3}$. When $Z(y_r)$ is PSD and the solution has rank three
with all blocks of positive determinant, the solve is certified
optimal for the original problem, and we report $\eta=10^{-12}$, the
numerical resolution of the eigenvalue solve. Otherwise, the
static $\eta$ uses the trace-restricted bound of Appendix~\ref{app:subopt}.
For the dynamic solver, each entry is one segment solve (its final
certificate), and the median and $n_{\mathrm{ok}}$ are over all segment solves.}
\label{tab:subopt_summary}
\footnotesize
\setlength{\tabcolsep}{4pt}
\begin{tabular}{lcccccc}
\toprule
\multirow{2}{*}{Dataset} & \multicolumn{3}{c}{Static} & \multicolumn{3}{c}{Dynamic (EDG)} \\
\cmidrule(lr){2-4}\cmidrule(lr){5-7}
 & scenes & median $\eta$ & $n_{\mathrm{ok}}$ & scenes & median $\eta$ & $n_{\mathrm{ok}}$ \\
\midrule
Sintel & 23 & $9.9\!\times\!10^{-12}$ & 23/23 & -- & -- & -- \\
TUM & 6 & $3.6\!\times\!10^{-11}$ & 6/6 & -- & -- & -- \\
Replica & 8 & $9.7\!\times\!10^{-10}$ & 8/8 & -- & -- & -- \\
BAL & 3 & $3.9\!\times\!10^{-5}$ & 3/3 & -- & -- & -- \\
IMC & 6 & $5.5\!\times\!10^{-12}$ & 6/6 & -- & -- & -- \\
DyCheck & 7 & $10^{-12}$ & 7/7 & 7 & $10^{-12}$ & 700/700 \\
NVIDIA & 7 & $10^{-12}$ & 7/7 & 7 & $3.5\!\times\!10^{-11}$ & 511/511 \\
Syn4D & 8 & $10^{-12}$ & 8/8 & 8 & $10^{-12}$ & 800/800 \\
\bottomrule
\end{tabular}
\end{table}

\begin{table}[!h]
\centering
\caption{Geometry accuracy on Syn4D per scene: Chamfer distance
($\times10^{-3}$) between the reconstructed
and the ground-truth dynamic object, and relative $L_1$ error of the
rendered depth in the training and novel views; lower is better. All
methods are scored with the same protocol: the Chamfer distance follows
the camera-based similarity alignment with one global scale per scene
chosen to minimize it, and the depth uses one median scale per scene,
fitted on the training views and reused for the novel views. SoM is Shape of Motion~\cite{wang25iccv-shapeofmotion}, and DA renders the Depth
Anything~3~\cite{lin25arxiv-da3} depth lifted with its cameras.}
\label{tab:syn4d-geometry}
\footnotesize
\setlength{\tabcolsep}{1.9pt}
\begin{tabular}{llccccccccc}
\toprule
Metric & Method & bear1 & bear2 & bear3 & bear4 & casa1 & casa2 & casa3 & casa4 & Mean \\
\midrule
\multirow{3}{*}{Chamfer}
& Ours & \textbf{6.1} & \textbf{5.4} & \textbf{6.4} & \textbf{4.6} & \textbf{11.2} & \textbf{9.2} & \textbf{7.6} & \textbf{8.6} & \textbf{7.4} \\

& MoSca & \underline{27.3} & \underline{29.1} & \underline{75.2} & \underline{45.6} & 112.0 & 125.1 & \underline{50.2} & \underline{60.4} & \underline{65.6} \\

& SoM & 34.4 & 214.7 & 90.8 & 172.4 & \underline{38.3} & \underline{33.3} & 69.3 & 2429.9 & 385.4 \\
\addlinespace[2pt]
\multirow{4}{*}{\shortstack[l]{Train\\depth}}
& Ours & \textbf{0.012} & \textbf{0.012} & \textbf{0.026} & \textbf{0.015} & \textbf{0.077} & \textbf{0.029} & \textbf{0.054} & \textbf{0.045} & \textbf{0.034} \\

& MoSca & \underline{0.031} & \underline{0.029} & \underline{0.035} & \underline{0.029} & \underline{0.123} & \underline{0.049} & \underline{0.081} & \underline{0.069} & \underline{0.056} \\

& SoM & 0.102 & 0.096 & 0.104 & 0.096 & 0.193 & 0.082 & 0.158 & 0.139 & 0.121 \\

& DA & 0.204 & 0.208 & 0.260 & 0.240 & 0.469 & 0.237 & 0.315 & 0.220 & 0.269 \\
\addlinespace[2pt]
\multirow{4}{*}{\shortstack[l]{Novel\\depth}}
& Ours & \textbf{0.050} & \textbf{0.094} & \textbf{0.094} & \textbf{0.085} & \textbf{0.101} & \textbf{0.038} & \textbf{0.055} & \textbf{0.053} & \textbf{0.071} \\

& MoSca & 0.296 & 0.329 & \underline{0.099} & \underline{0.292} & \underline{0.171} & 0.182 & \underline{0.121} & \underline{0.087} & \underline{0.197} \\

& SoM & 0.321 & 0.322 & 0.273 & 0.334 & 0.253 & \underline{0.143} & 0.188 & 0.210 & 0.255 \\

& DA & \underline{0.210} & \underline{0.286} & 0.368 & 0.296 & 0.458 & 0.255 & 0.357 & 0.262 & 0.312 \\
\bottomrule
\end{tabular}
\end{table}

%% file: sections/app-proofs.tex
\section{Proofs}
\label{app:proofs}

\subsection{Proof of Proposition~\ref{prop:rotation-only}}
\label{app:rotation-only}

\begin{proof}
Let $\mathbf{e}_t\in\mathbb{R}^{T}$ and
$\bar{\mathbf{e}}_i\in\mathbb{R}^{N_s}$ denote standard basis
vectors. All vectors are columns; collect the unknown translations
and landmarks as the \emph{columns} of the matrix
\begin{equation}
Y=
\begin{bmatrix}
\mathbf{c}_1 & \cdots & \mathbf{c}_T &
\mathbf{x}^{s}_1 & \cdots & \mathbf{x}^{s}_{N_s}
\end{bmatrix}
\in\mathbb{R}^{3\times(T+N_s)}.
\end{equation}
For each observation $(i,t)\in\mathcal{E}_s$ define
\begin{equation}
\mathbf{a}_{it}=\mathbf{e}_t\otimes\tilde{\mathbf{u}}^{s}_{it}
\in\mathbb{R}^{3T},
\qquad
\mathbf{b}_{it}=
\begin{bmatrix}
\mathbf{e}_t\\
-\bar{\mathbf{e}}_i
\end{bmatrix}
\in\mathbb{R}^{T+N_s},
\end{equation}
so that, with $U=[s_1C_1\;\cdots\;s_TC_T]$, each residual of
Eq.~\eqref{eq:sba} reads
$C_t(s_t\tilde{\mathbf{u}}^{s}_{it})+\mathbf{c}_t-\mathbf{x}^{s}_i
=U\mathbf{a}_{it}+Y\mathbf{b}_{it}$. The objective of
Eq.~\eqref{eq:sba} is therefore the quadratic form
\begin{equation}
f(U,Y)=
\operatorname{tr}(Q_1U^\top U)
+2\operatorname{tr}(V^\top U^\top Y)
+\operatorname{tr}(L_{\mathcal{G}}Y^\top Y),
\label{eq:app-quadratic}
\end{equation}
with
\begin{align}
Q_1&=\textstyle\sum_{(i,t)\in\mathcal{E}_s}
\omega_{it}\,\mathbf{a}_{it}\mathbf{a}_{it}^\top
\in\mathbb{S}^{3T},
\label{eq:app-Q1}\\
V&=\textstyle\sum_{(i,t)\in\mathcal{E}_s}
\omega_{it}\,\mathbf{a}_{it}\mathbf{b}_{it}^\top
\in\mathbb{R}^{3T\times(T+N_s)},
\label{eq:app-V}\\
L_{\mathcal{G}}&=\textstyle\sum_{(i,t)\in\mathcal{E}_s}
\omega_{it}\,\mathbf{b}_{it}\mathbf{b}_{it}^\top
\in\mathbb{S}^{T+N_s}.
\label{eq:app-L}
\end{align}
By construction, $L_{\mathcal{G}}$ is the weighted Laplacian of the
bipartite view graph $\mathcal{G}$ on the $T+N_s$ vertices
$\{\text{frames}\}\cup\{\text{landmarks}\}$ with edge weights
$\omega_{it}$ (cf.~\cite{han25rss-xm}): $\mathbf{b}_{it}^\top
\mathbf{1}=0$ implies $L_{\mathcal{G}}\mathbf{1}=\mathbf{0}$, and
when $\mathcal{G}$ is connected the nullspace of $L_{\mathcal{G}}$
is exactly $\operatorname{span}(\mathbf{1})$. Likewise,
$V\mathbf{1}=\sum\omega_{it}\mathbf{a}_{it}
(\mathbf{b}_{it}^\top\mathbf{1})=\mathbf{0}$.

For fixed $U$, $f$ is a convex quadratic in $Y$, bounded below by
zero, so its minimizers are the solutions of the stationarity
condition $YL_{\mathcal{G}}=-UV$. This linear system is consistent:
$V\mathbf{1}=\mathbf{0}$ places the rows of $-UV$ in the range of
$L_{\mathcal{G}}$. Its minimum-norm solution is
\begin{equation}
Y^\star=-UVL_{\mathcal{G}}^{+},
\label{eq:app-ystar}
\end{equation}
and every other solution differs by $\mathbf{v}\mathbf{1}^\top$.
Substituting~\eqref{eq:app-ystar} into~\eqref{eq:app-quadratic} and
using $L_{\mathcal{G}}^{+}L_{\mathcal{G}}L_{\mathcal{G}}^{+}
=L_{\mathcal{G}}^{+}$ gives
\begin{equation}
\min_{Y}f(U,Y)=
\operatorname{tr}\!\left(QU^\top U\right),
\qquad
Q=Q_1-VL_{\mathcal{G}}^{+}V^\top,
\label{eq:schur-Q}
\end{equation}
the generalized Schur complement of $L_{\mathcal{G}}$ in
$\sum\omega_{it}[\mathbf{a}_{it};\mathbf{b}_{it}]
[\mathbf{a}_{it};\mathbf{b}_{it}]^\top\succeq0$; since the columns
of $V^\top$ lie in $\operatorname{range}(L_{\mathcal{G}})$,
$Q\succeq0$. Minimizing both
sides of~\eqref{eq:schur-Q} over $\{s_t>0,\;C_t\in\mathrm{SO}(3)\}$
proves the claimed equivalence, and~\eqref{eq:app-ystar} recovers
the optimal translations and landmarks from $U^\star$.\looseness=-1
\\
\end{proof}

\subsection{Proof of Proposition~\ref{prop:qcqp}}
\label{app:qcqp}

Writing a generic $3\times3$ matrix as
$\bar C=[\mathbf{a}_1,\mathbf{a}_2,\mathbf{a}_3]$, membership in the
set~\eqref{eq:so3-scaled} of scaled orthogonal matrices is described by the
quadratic equalities
\begin{align}
\mathbf{a}_1^\top \mathbf{a}_1
&=
\mathbf{a}_2^\top \mathbf{a}_2
=
\mathbf{a}_3^\top \mathbf{a}_3,
\label{eq:sO-norm}\\
\mathbf{a}_1^\top \mathbf{a}_2
&=
\mathbf{a}_2^\top \mathbf{a}_3
=
\mathbf{a}_3^\top \mathbf{a}_1
=
0,
\label{eq:sO-orth}
\end{align}
which, together with the quadratic sphere constraint
$\norm{U}_F^2=3T$, are the constraints of~\eqref{eq:qcqp}.

\begin{proof}
We first characterize the feasible set. If
$\bar C=[\mathbf{a}_1,\mathbf{a}_2,\mathbf{a}_3]$ satisfies the
quadratic equalities~\eqref{eq:sO-norm}--\eqref{eq:sO-orth}, then
$\bar C^\top\bar C=\sigma^{2}I_3$ with
$\sigma^{2}=\mathbf{a}_1^\top\mathbf{a}_1\geq0$. If $\sigma>0$,
then $\bar C/\sigma\in\mathrm{O}(3)$ and
$\bar C=\sigma R$; if $\sigma=0$, then $\bar C=0$. The quadratic
constraints therefore describe exactly the set~\eqref{eq:so3-scaled},
including the zero matrix, and the feasible set
of~\eqref{eq:qcqp} is
$\{U:\bar C_t\in\mathrm{sO}(3),\;\norm{U}_F^2=3T\}$.
This set contains the feasible set of~\eqref{eq:sro}
under the sphere constraint: for $s_t>0$ and
$C_t\in\mathrm{SO}(3)$, each block $\bar C_t=s_tC_t$ lies in
$\mathrm{sO}(3)$, and
$\norm{U}_F^2=\sum_{t}3s_t^2=3T$ holds if and only if
$\sum_{t}s_t^2=T$. The QCQP thus
relaxes~\eqref{eq:sro} in exactly two ways: reflections
are admitted ($\mathrm{SO}(3)$ is relaxed to $\mathrm{O}(3)$), and
individual blocks may vanish. Since the objectives coincide on the
common feasible set,
\begin{equation}
\rho^\star_{\mathrm{QCQP}}
=\operatorname{tr}\!\left(Q(U^\star)^\top U^\star\right)
\leq\rho^\star_{\mathrm{SRO}}.
\label{eq:app-qcqp-relax}
\end{equation}

Now let $U^\star$ be a global optimizer of~\eqref{eq:qcqp} with
$\det\bar C_t^\star>0$ for all $t$. The determinant condition rules
out both relaxations at once: every block is nonzero, so
$\bar C_t^\star=s_t^\star C_t^\star$ with
$s_t^\star=\norm{\bar C_t^\star}_F/\sqrt{3}>0$ and
$C_t^\star\in\mathrm{O}(3)$, and
\begin{equation}
\det C_t^\star=\frac{\det\bar C_t^\star}{(s_t^\star)^{3}}>0,
\end{equation}
i.e., $C_t^\star\in\mathrm{SO}(3)$. Moreover,
$\sum_{t}(s_t^\star)^{2}=\norm{U^\star}_F^{2}/3=T$, so
$\{s_t^\star,C_t^\star\}_{t=1}^{T}$ is feasible
for~\eqref{eq:sro} under the sphere constraint, which
gives $\operatorname{tr}(Q(U^\star)^\top U^\star)\geq\rho^\star_{\mathrm{SRO}}$.
Combined with~\eqref{eq:app-qcqp-relax}, equality holds, and
$U^\star$ is a global minimizer of~\eqref{eq:sro}.
Finally, since $(GU)^\top(GU)=U^\top U$,
$GU^\star$ is equally optimal for every $G\in\mathrm{O}(3)$, and
by~\eqref{eq:app-ystar} the recovered translations and landmarks
rotate with it; the normalization
$U^\star\gets(C_1^\star)^\top U^\star$ selects the representative
that aligns the first camera with the identity.
\end{proof}

\subsection{Proof of Proposition~\ref{prop:sdp}}
\label{app:sdp}

\begin{proof}
\emph{Relaxation.} Let $U=[\bar C_1,\dots,\bar C_T]$ be feasible
for~\eqref{eq:qcqp} and set $X=U^\top U$. Then each diagonal block
is $X_{tt}=\bar C_t^\top\bar C_t=\alpha_tI_3$
by~\eqref{eq:so3-scaled}; and
$\operatorname{tr}(X)=\norm{U}_F^2=3T$. Hence $X$ is feasible
for~\eqref{eq:sdp} with the same objective value
$\langle Q,X\rangle=\operatorname{tr}(QU^\top U)$, and minimizing
over $U$ gives $\rho^\star_{\mathrm{SDP}}\leq\rho^\star_{\mathrm{QCQP}}$.

\emph{Tightness.} Let $X^\star$ be a global minimizer
of~\eqref{eq:sdp} with $\operatorname{rank}(X^\star)=3$, and write
its eigendecomposition as
$X^\star=\sum_{k=1}^{3}\lambda_k\mathbf{v}_k\mathbf{v}_k^\top$ with
$\lambda_k>0$. The matrix $U^\star\in\mathbb{R}^{3\times3T}$ with
rows $\sqrt{\lambda_k}\,\mathbf{v}_k^\top$ satisfies
$X^\star=(U^\star)^\top U^\star$. Partitioning
$U^\star=[\bar C_1^\star,\dots,\bar C_T^\star]$ into $3\times3$
blocks, the block constraints give
$(\bar C_t^\star)^\top\bar C_t^\star=X^\star_{tt}=\alpha_tI_3$, so
$\bar C_t^\star\in\mathrm{sO}(3)$ by~\eqref{eq:so3-scaled}, and the
trace constraint gives $\norm{U^\star}_F^2=\operatorname{tr}(X^\star)=3T$.
Thus $U^\star$ is feasible for~\eqref{eq:qcqp}, and
\begin{equation}
\rho^\star_{\mathrm{QCQP}}
\leq\operatorname{tr}\!\left(Q(U^\star)^\top U^\star\right)
=\langle Q,X^\star\rangle
=\rho^\star_{\mathrm{SDP}}
\leq\rho^\star_{\mathrm{QCQP}},
\end{equation}
so equality holds and $U^\star$ is a global optimizer
of~\eqref{eq:qcqp}.

\end{proof}

\subsection{Proof of Proposition~\ref{prop:bm-manifold}}
\label{app:bm-manifold}

\begin{proof}
Let $U$ be feasible for~\eqref{eq:bm-block}. Each block satisfies
$U_t^\top U_t=\alpha_tI_3$ with $\alpha_t\geq0$; set
$s_t=\sqrt{\alpha_t}$. If $s_t>0$, then $Y_t=U_t/s_t$ satisfies
$Y_t^\top Y_t=I_3$, i.e., $Y_t\in\mathrm{St}(r,3)$; if $s_t=0$, then
$U_t=0$ and any $Y_t\in\mathrm{St}(r,3)$ gives $U_t=s_tY_t$. The trace
constraint $\sum_t\alpha_t=T$ reads $\norm{\mathbf{s}}_2^2=T$, so
$(\mathbf{s},Y_1,\dots,Y_T)\in\mathcal{M}$. Conversely, every point
of $\mathcal{M}$ gives blocks $U_t=s_tY_t$ with
$U_t^\top U_t=s_t^2I_3$ and $\sum_ts_t^2=T$, which is feasible
for~\eqref{eq:bm-block} with $\alpha_t=s_t^2$. The map
$(\mathbf{s},Y_1,\dots,Y_T)\mapsto U$ is therefore onto the feasible
set of~\eqref{eq:bm-block}, and the two problems share the objective
$\operatorname{tr}(QU^\top U)$, so they are equivalent.
\end{proof}

\subsection{Proof of Theorem~\ref{thm:certificate}}
\label{app:licq}

\begin{proof}
We first make the constraint matrices $\{A_i\}_{i=1}^{m}$
of~\eqref{eq:sdp-standard} explicit. For each frame $t$, the block
constraint $X_{tt}=\alpha_tI_3$ for some $\alpha_t$ is equivalent to
the five homogeneous equations
$\langle A_{5(t-1)+\ell},X\rangle=0$, $\ell=1,\dots,5$, where
$A_{5(t-1)+\ell}\in\mathbb{S}^{3T}$ is zero except for its $t$-th
diagonal $3\times3$ block, which equals
\begin{equation}
\begin{gathered}
B^1=\left[\begin{smallmatrix}1&0&0\\0&-1&0\\0&0&0\end{smallmatrix}\right],\;
B^2=\left[\begin{smallmatrix}0&0&0\\0&1&0\\0&0&-1\end{smallmatrix}\right],\;
B^3=\left[\begin{smallmatrix}0&1&0\\1&0&0\\0&0&0\end{smallmatrix}\right],\\
B^4=\left[\begin{smallmatrix}0&0&1\\0&0&0\\1&0&0\end{smallmatrix}\right],\;
B^5=\left[\begin{smallmatrix}0&0&0\\0&0&1\\0&1&0\end{smallmatrix}\right].
\end{gathered}
\label{eq:app-basis}
\end{equation}
The $\{B^\ell\}_{\ell=1}^{5}$ form a basis of the traceless
symmetric $3\times3$ matrices. The trace
constraint $\operatorname{tr}(X)=3T$ of~\eqref{eq:sdp} corresponds to $A_m=I_{3T}$
with $b_m=3T$.

\emph{LICQ.} The gradient of the $i$-th constraint of the BM
problem at $U$ is $2A_iU^\top$. Suppose
$\sum_{i=1}^{m}\gamma_iA_iU^\top=0$ for some coefficients
$\gamma\in\mathbb{R}^{m}$, evaluated at $U=U_r^\star$. The matrix
$A_{5(t-1)+\ell}U^\top$ is supported on rows $3t-2,\dots,3t$ only,
while $A_mU^\top=U^\top$ contributes $U_t^\top$ to those same rows;
collecting the rows of block $t$ therefore gives
\begin{equation}
\left(\sum_{\ell=1}^{5}\gamma_{t\ell}B^\ell+\gamma_mI_3\right)
U_t^\top=0,
\qquad t=1,\dots,T,
\label{eq:app-licq-block}
\end{equation}
where $\gamma_{t\ell}=\gamma_{5(t-1)+\ell}$. Since $\alpha_t>0$,
the block $U_t$ has full column rank; multiplying
\eqref{eq:app-licq-block} on the right by $U_t$ and using
$U_t^\top U_t=\alpha_tI_3$ yields
$\sum_{\ell}\gamma_{t\ell}B^\ell+\gamma_mI_3=0$. Taking the trace
gives $3\gamma_m=0$ because every $B^\ell$ is traceless, so
$\gamma_m=0$; linear independence of $\{B^\ell\}$ then forces
$\gamma_{t\ell}=0$ for all $t,\ell$. Hence the constraint gradients
$\{2A_iU^\top\}_{i=1}^{m}$ are linearly independent, and LICQ holds
at $U_r^\star$. 

\emph{First-order condition.} By LICQ, there exists a unique
Lagrange multiplier $y_r\in\mathbb{R}^{m}$ at which the Lagrangian
of~\eqref{eq:bm-block} is stationary:
\begin{equation}
Q(U_r^\star)^\top-\sum_{i=1}^{m}(y_r)_iA_i(U_r^\star)^\top
=Z(y_r)(U_r^\star)^\top=0,
\end{equation}
which is exactly~\eqref{eq:first-order}.

\emph{Global optimality.} Suppose in addition $Z(y_r)\succeq0$.
Then $y_r$ is feasible for the dual~\eqref{eq:sdp-dual}, and
$X^\star=(U_r^\star)^\top U_r^\star$ is feasible for the
primal~\eqref{eq:sdp-standard}. Their objective values coincide:
\begin{equation}
\begin{split}
\operatorname{tr}(QX^\star)-\sum_{i=1}^{m}b_i(y_r)_i
&=\langle Z(y_r),X^\star\rangle\\
&=\operatorname{tr}\!\left(Z(y_r)(U_r^\star)^\top U_r^\star\right)
=0,
\end{split}
\end{equation}
using primal feasibility in the first equality
and~\eqref{eq:first-order} in the last. The duality gap is zero, so
$X^\star$ and $y_r$ are optimal for the primal--dual pair, and
$U_r^\star$ attains
$\rho^\star_{\mathrm{SDP}}\leq\rho^\star_{\mathrm{BM},r}$, hence it
is a global optimizer
of~\eqref{eq:bm-block}~\cite{burer03mp-nonlinear,Yang24book-sdp}.
\end{proof}

\subsection{Proof of Theorem~\ref{thm:descent}}
\label{app:descent}

\begin{proof}
See~\cite{burer03mp-nonlinear} or~\cite{Yang24book-sdp}.
\end{proof}

\subsection{Proof of Theorem~\ref{thm:global}}
\label{app:global}

\begin{proof}
In the worst case $r=3T$, every local optimizer
of~\eqref{eq:bm-block} corresponds to a globally optimal solution
of the SDP~\cite[Corollary~8]{journee10siamopt-lowrank}: if the local
optimizer is rank-deficient, global optimality follows
from~\cite[Proposition~4]{burer03mp-nonlinear}; if it is full-rank,
then~\eqref{eq:first-order} forces $Z(y_r)=0$, which is positive
semidefinite and again certifies global optimality.
\end{proof}

\subsection{Proof of Proposition~\ref{prop:dyn-rotation-only}}
\label{app:dyn-rotation-only}

\begin{proof}
Let $\Omega=\{(j,t):j=1,\dots,J,\;t\neq t_0\}$ be the free
node-frame pairs of~\eqref{eq:dyn}, $n=|\Omega|=J(T-1)$, and fix an
ordering of $\Omega$, the one used in $U$. Let $\mathbf{e}_{jt}\in\mathbb{R}^{n}$
denote the standard basis vector of the pair $(j,t)\in\Omega$, set
$\mathbf{e}_{jt_0}=\mathbf{0}$, and let $\delta_{tt_0}$ be the
Kronecker delta. Collect the unknown node translations as the \emph{columns}
of
\begin{equation}
Y=
\begin{bmatrix}
\mathbf{p}_{j,t}
\end{bmatrix}_{(j,t)\in\Omega}
\in\mathbb{R}^{3\times n},
\end{equation}
in the same order as $U$. Then, for every node $j$, frame $t$, and
vector $\mathbf{v}\in\mathbb{R}^{3}$, the rest-frame
convention~\eqref{eq:edg-rest} gives
\begin{equation}
R_{j,t}\mathbf{v}
=U(\mathbf{e}_{jt}\otimes\mathbf{v})+\delta_{tt_0}\mathbf{v},
\qquad
\mathbf{p}_{j,t}=Y\mathbf{e}_{jt},
\label{eq:app-dyn-select}
\end{equation}
so the fixed rest-frame blocks contribute constants only.

\textbf{Residuals.} By~\eqref{eq:app-dyn-select}, every residual
$\ell$ of~\eqref{eq:dyn} has the form
\begin{equation}
\mathbf{r}_\ell=U\mathbf{a}_\ell+Y\mathbf{b}_\ell+\mathbf{c}_\ell,
\qquad
\mathbf{a}_\ell\in\mathbb{R}^{3n},\;
\mathbf{b}_\ell\in\mathbb{R}^{n},\;
\mathbf{c}_\ell\in\mathbb{R}^{3},
\label{eq:app-dyn-residual}
\end{equation}
with weight $\lambda_\ell$ equal to the coefficient of its term
in~\eqref{eq:dyn}. Using $\sum_{j\in\mathcal{N}(i)}w_{ij}=1$, the
three terms give the following.
\begin{itemize}
\item \emph{Tracking}, one residual per $(i,t)\in\mathcal{E}_d$
with weight $\lambda_{\mathrm{track}}$:
\begin{equation}
\begin{split}
\mathbf{a}&=\textstyle\sum_{j\in\mathcal{N}(i)}w_{ij}\,
\mathbf{e}_{jt}\otimes(\mathbf{x}^{d}_i-\mathbf{g}_j),\\
\mathbf{b}&=\textstyle\sum_{j\in\mathcal{N}(i)}w_{ij}\,
\mathbf{e}_{jt},\\
\mathbf{c}&=\bar{\mathbf{g}}_i
+\delta_{tt_0}(\mathbf{x}^{d}_i-\bar{\mathbf{g}}_i)
-\mathbf{y}^{d}_{it},
\quad
\bar{\mathbf{g}}_i=\textstyle\sum_{j\in\mathcal{N}(i)}w_{ij}\,\mathbf{g}_j.
\end{split}
\end{equation}
\item \emph{ARAP}, one residual per edge $(j,k)\in
\mathcal{E}_{\mathrm{EDG}}$ and frame $t\neq t_0$ with weight
$\lambda_{\mathrm{ARAP}}$ (at $t=t_0$ the residual vanishes
identically):
\begin{equation}
\mathbf{a}=\mathbf{e}_{jt}\otimes(\mathbf{g}_k-\mathbf{g}_j),\qquad
\mathbf{b}=\mathbf{e}_{jt}-\mathbf{e}_{kt},\qquad
\mathbf{c}=\mathbf{g}_j-\mathbf{g}_k.
\end{equation}
\item \emph{Temporal smoothness}, for every node $j$ and
$t=1,\dots,T-1$ with weight $\lambda_{\mathrm{temp}}$: writing
\[
\norm{R_{j,t}-R_{j,t+1}}_F^2=\sum_{m=1}^{3}
\norm{(R_{j,t}-R_{j,t+1})\boldsymbol{\epsilon}_m}^2
\]
with
$\boldsymbol{\epsilon}_m$ the standard basis of $\mathbb{R}^{3}$,
three rotation residuals ($m=1,2,3$) and one translation residual,
\begin{equation}
\begin{split}
&\mathbf{a}=(\mathbf{e}_{jt}-\mathbf{e}_{j,t+1})\otimes
\boldsymbol{\epsilon}_m,\;\;
\mathbf{b}=\mathbf{0},\\
&\mathbf{c}=(\delta_{tt_0}-\delta_{t+1,t_0})\boldsymbol{\epsilon}_m,\\
&\mathbf{a}=\mathbf{0},\;\;
\mathbf{b}=\mathbf{e}_{jt}-\mathbf{e}_{j,t+1},\;\;
\mathbf{c}=\mathbf{0},
\end{split}
\end{equation}
\end{itemize}
The node term of the implementation
(Appendix~\ref{app:impl-node}) has the same form, with
$\mathbf{a}=\mathbf{0}$, $\mathbf{b}=\mathbf{e}_{jt}$, and
$\mathbf{c}=\mathbf{g}_j-\mathbf{y}^{d}_{jt}$, so the reduction below
applies unchanged when it is included.

\textbf{Quadratic form.} Summing $\lambda_\ell\norm{\mathbf{r}_\ell}^2$
over all residuals, the objective of~\eqref{eq:dyn} is
\begin{equation}
\begin{split}
f(U,Y)={}&
\operatorname{tr}(Q_1U^\top U)
+2\operatorname{tr}(W^\top U^\top Y)
+\operatorname{tr}(LY^\top Y)\\
&+2\langle P_1,U\rangle
+2\langle P_Y,Y\rangle
+c_1,
\end{split}
\label{eq:app-dyn-quadratic}
\end{equation}
with
\begin{align}
Q_1&=\textstyle\sum_\ell\lambda_\ell\,\mathbf{a}_\ell\mathbf{a}_\ell^\top
\in\mathbb{S}^{3n},
&
W&=\textstyle\sum_\ell\lambda_\ell\,\mathbf{a}_\ell\mathbf{b}_\ell^\top
\in\mathbb{R}^{3n\times n},
\label{eq:app-dyn-Q1W}\\
L&=\textstyle\sum_\ell\lambda_\ell\,\mathbf{b}_\ell\mathbf{b}_\ell^\top
\in\mathbb{S}^{n},
&
P_1&=\textstyle\sum_\ell\lambda_\ell\,\mathbf{c}_\ell\mathbf{a}_\ell^\top
\in\mathbb{R}^{3\times3n},
\label{eq:app-dyn-LP1}\\
P_Y&=\textstyle\sum_\ell\lambda_\ell\,\mathbf{c}_\ell\mathbf{b}_\ell^\top
\in\mathbb{R}^{3\times n},
&
c_1&=\textstyle\sum_\ell\lambda_\ell\,\norm{\mathbf{c}_\ell}^2.
\label{eq:app-dyn-PYc}
\end{align}

\textbf{Elimination.} For fixed $U$, $f$ is a convex quadratic in
$Y$, bounded below by zero, so its minimizers solve the stationarity
condition $YL=-(UW+P_Y)$. This system is consistent: the rows of
$UW+P_Y=\sum_\ell\lambda_\ell(U\mathbf{a}_\ell+\mathbf{c}_\ell)\mathbf{b}_\ell^\top$
lie in $\operatorname{span}\{\mathbf{b}_\ell\}=\operatorname{range}(L)$.
Its minimum-norm solution is
\begin{equation}
Y^\star=-(UW+P_Y)L^{+},
\label{eq:app-dyn-ystar}
\end{equation}
which is unique when $L\succ0$, e.g., whenever $\lambda_{\mathrm{temp}}>0$, since every node's temporal chain then reaches the fixed rest frame. Substituting~\eqref{eq:app-dyn-ystar}
into~\eqref{eq:app-dyn-quadratic} and using
$L^{+}LL^{+}=L^{+}$ gives
$\min_Yf(U,Y)=\operatorname{tr}(QU^\top U)+2\langle P,U\rangle+c_0$
with
\begin{equation}
\begin{split}
Q&=Q_1-WL^{+}W^\top,\\
P&=P_1-P_YL^{+}W^\top,\\
c_0&=c_1-\operatorname{tr}\!\left(P_YL^{+}P_Y^\top\right).
\end{split}
\label{eq:app-dyn-QP}
\end{equation}
The matrix $Q$ is the generalized Schur complement of $L$ in
$\sum_\ell\lambda_\ell[\mathbf{a}_\ell;\mathbf{b}_\ell]
[\mathbf{a}_\ell;\mathbf{b}_\ell]^\top\succeq0$, and the columns of
$W^\top$ lie in $\operatorname{range}(L)$, so $Q\succeq0$.
Minimizing both sides over $\{R_{j,t}\in\mathrm{SO}(3)\}_{(j,t)\in
\Omega}$ proves the claimed equivalence,
and~\eqref{eq:app-dyn-ystar} recovers the optimal node translations
from $U^\star$.
\end{proof}

\subsection{Proof of Proposition~\ref{prop:dyn-homog}}
\label{app:dyn-homog}

\begin{proof}
Since $V^\top V=I_3$,
\begin{equation}
U_h^\top U_h=
\begin{bmatrix}
I_3 & V^\top U\\
U^\top V & U^\top U
\end{bmatrix},
\end{equation}
and therefore
\begin{equation}
\begin{split}
\operatorname{tr}\!\left(Q_hU_h^\top U_h\right)
&=\tfrac{c_0}{3}\operatorname{tr}(I_3)
+\operatorname{tr}\!\left(PU^\top V\right)\\
&\quad+\operatorname{tr}\!\left(P^\top V^\top U\right)
+\operatorname{tr}\!\left(QU^\top U\right)\\
&=\operatorname{tr}\!\left(QU^\top U\right)
+2\langle P,V^\top U\rangle+c_0.
\end{split}
\label{eq:app-dyn-homog-identity}
\end{equation}
Let $\widetilde U=V^\top U$. Orthogonality of $V$ gives
$U^\top U=\widetilde U^\top\widetilde U$, so
by~\eqref{eq:app-dyn-homog-identity}
\begin{equation}
\operatorname{tr}\!\left(Q_hU_h^\top U_h\right)
=\operatorname{tr}\!\left(Q\widetilde U^\top\widetilde U\right)
+2\langle P,\widetilde U\rangle+c_0,
\end{equation}
which is the objective of~\eqref{eq:dyn-ro} evaluated at
$\widetilde U$. The blocks of $\widetilde U$ are $V^\top R_{j,t}$,
which lie in $\mathrm{SO}(3)$ whenever $V$ and $R_{j,t}$ do, so
$\widetilde U$ is feasible for~\eqref{eq:dyn-ro}. Hence every
feasible $(V,U)$ of~\eqref{eq:dyn-h} attains the objective value of
a feasible point of~\eqref{eq:dyn-ro}, and
$\rho^\star_{\mathrm{DYN\text{-}H}}\geq
\rho^\star_{\mathrm{DYN\text{-}RO}}$. Conversely, every feasible
$\widetilde U$ of~\eqref{eq:dyn-ro} is attained by the feasible pair
$(V,U)=(I_3,\widetilde U)$, so $\rho^\star_{\mathrm{DYN\text{-}H}}
\leq\rho^\star_{\mathrm{DYN\text{-}RO}}$. The two optimal values
therefore coincide, and the correspondence of
minimizers follows from the equality of objective values along the
map $(V,U)\mapsto V^\top U$. The same argument applies verbatim with
$\mathrm{O}(3)$ in place of $\mathrm{SO}(3)$.
\end{proof}

\subsection{Proof of Proposition~\ref{prop:dyn-sdp-bm}}
\label{app:dyn-sdp-bm}

\begin{proof}
Identical to Propositions~\ref{prop:sdp}--\ref{prop:bm-manifold} and
Theorems~\ref{thm:certificate}--\ref{thm:global}, with every
block constraint $X_{ii}=\alpha_iI_3$ fixed to $X_{ii}=I_3$ and the
trace constraint removed: $X=I$ remains strictly feasible, so strong duality holds,
and each block contributes six linear equalities, so LICQ
holds as in Appendix~\ref{app:licq}.
\end{proof}

%% file: sections/app-subopt.tex
\section{Suboptimality Gap for the Sphere-Constrained SDP}
\label{app:subopt}

Algorithm~\ref{alg:staircase} is run with finite gradient
tolerances, so the returned $Z(y_r)$ may have a slightly negative
minimum eigenvalue even when the solution is essentially optimal.
Adapting the suboptimality analysis of XM~\cite{han25rss-xm} to the
sphere-constrained relaxation~\eqref{eq:sdp-standard}, the
following bound quantifies the resulting optimality gap.

\begin{theorem}[Suboptimality Bound, adapted from~\cite{han25rss-xm}]
\label{thm:subopt}
Let $\hat X$ be any feasible point of~\eqref{eq:sdp-standard} with
objective value $\hat\rho=\operatorname{tr}(Q\hat X)$, let
$\mathbf{b}=(b_1,\dots,b_m)$ collect the right-hand sides
of~\eqref{eq:sdp-standard-linear}, let $Z(y)$ be the dual slack
of~\eqref{eq:dual-cert}, and let $y\in\mathbb{R}^{m}$ be arbitrary.
Then
\begin{equation}
\hat\rho
\;\geq\;
\rho^\star_{\mathrm{SDP}}
\;\geq\;
\mathbf{b}^\top y+3T\,\lambda_{\min}\!\left(Z(y)\right).
\label{eq:subopt-bound}
\end{equation}
\end{theorem}
\begin{proof}
The first inequality holds because $\hat X$ is feasible. For the
second, let $X$ be any feasible point
of~\eqref{eq:sdp-standard}. Using
$\langle A_i,X\rangle=b_i$,
\begin{equation}
\operatorname{tr}(QX)
=\langle Z(y),X\rangle+\sum_{i=1}^{m}y_i\langle A_i,X\rangle
=\langle Z(y),X\rangle+\mathbf{b}^\top y.
\end{equation}
Since $X\succeq0$, we have
$\langle Z(y),X\rangle\geq
\lambda_{\min}(Z(y))\operatorname{tr}(X)$, and the trace
constraint fixes $\operatorname{tr}(X)=3T$; hence
$\operatorname{tr}(QX)\geq \mathbf{b}^\top y
+3T\lambda_{\min}(Z(y))$. Minimizing over all feasible $X$ yields
the claim.
\end{proof}

In the static stage we apply~\eqref{eq:subopt-bound} with
$\hat X=\hat U^\top\hat U$, the rounded solution of
Appendix~\ref{app:rounding}, and $y=y_r$
from the certification step~\eqref{eq:first-order}; the resulting
relative gap $\eta$ is defined in~\eqref{eq:static-gap}
(Section~\ref{sec:sphere}). 
The same argument applies to the dynamic
relaxation~\eqref{eq:dyn-sdp}, whose block constraints $X_{ii}=I_3$
fix $\operatorname{tr}(X)=3(n+1)$; there the bound reads
$\mathbf{b}^\top y+3(n+1)\lambda_{\min}(Z(y))$.

\textbf{Partial sphere constraint.} The partial sphere of
Section~\ref{sec:pipe-static} fixes only
$\sum_{t\in\mathcal{S}}\operatorname{tr}(X_{tt})$, so
$\operatorname{tr}(X)$ is no longer constant. Restricting
the partial-sphere variant of~\eqref{eq:sdp-standard}
to $\operatorname{tr}(X)\leq\tau$, the same proof gives
$\rho^\star_\tau\geq\mathbf{b}^\top y+\tau\min\{0,\lambda_{\min}(Z(y))\}$,
and without the restriction when $Z(y)\succeq0$. In our implementation
we take $\tau=2\operatorname{tr}(\hat X)$, which bounds the
suboptimality of $\hat X$ among feasible points whose trace does not
exceed twice its own.

\subsection{Rounding to the Scaled-Rotation Problem}
\label{app:rounding}

Let $U_r^\star\in\mathbb{R}^{r\times3T}$ be the output of
Algorithm~\ref{alg:staircase}. We round it to a feasible point
$\hat U=[s_1C_1\;\cdots\;s_TC_T]$ of~\eqref{eq:sro} under the sphere
constraint $\norm{\hat U}_F^2=3T$ in three steps.
\begin{enumerate}
\item \emph{Rank-three factor.} Let $(\Lambda_3,V_3)$ be the three
largest eigenpairs of $(U_r^\star)^\top U_r^\star$ and set
$\tilde U=\Lambda_3^{1/2}V_3^\top\in\mathbb{R}^{3\times3T}$, with
blocks $\tilde U_t\in\mathbb{R}^{3\times3}$. 
\item \emph{Projection.} Each block is projected onto the scaled
rotations: with the singular value decomposition
$\tilde U_t=A_t\Sigma_tB_t^\top$, we set
$C_t=A_t\operatorname{diag}\!\big(1,1,\det(A_tB_t^\top)\big)B_t^\top
\in\mathrm{SO}(3)$ and
$s_t=\langle\tilde U_t,C_t\rangle/3$.
\item \emph{Sphere.} We rescale all scales by the common factor
$\sqrt{T}/\norm{\mathbf{s}}_2$.
\end{enumerate}
Under the partial sphere constraint, step~3 uses the factor
$\sqrt{|\mathcal{S}|}/\norm{\mathbf{s}_{\mathcal{S}}}_2$, where
$\mathbf{s}_{\mathcal{S}}$ collects the scales of the frames
in~$\mathcal{S}$. This enforces
$\sum_{t\in\mathcal{S}}s_t^2=|\mathcal{S}|$.

%% file: sections/app-datasets.tex
\section{Dataset Details}
\label{app:datasets}

\textbf{IMC.} Six scenes of the IMC PhotoTourism
benchmark~\cite{Jin20ijcv-imc2023}---gate\_00, gate\_01, rome\_00,
rome\_04, temple\_00, and temple\_01---each an unordered collection
of internet photographs of a static landmark, subsampled from the
original collection to about $200$ images per scene.

\textbf{Replica.} Eight synthetic indoor sequences rendered from the
Replica dataset~\cite{straub19arxiv-replica} as in
NICE-SLAM~\cite{Zhu22cvpr-niceslam}: office1, office2, room1, and
room2, two sequences each. Each sequence is subsampled from the
original $2000$ frames to $100$ frames. The scenes are static.\looseness=-1

\textbf{TUM.} Six sequences from the TUM RGB-D
benchmark~\cite{sturm12iros-tumdataset}, each subsampled from the
original sequence to $100$ frames. The sequences contain moving
objects; only camera accuracy is evaluated.

\textbf{BAL.} Three problems of the Bundle Adjustment in the Large
dataset~\cite{agarwal10eccv-bal}, BAL-93, BAL-394, and BAL-1936,
used only for the solver-configuration study.

\textbf{Sintel.} Twenty-three synthetic sequences of the MPI Sintel
dataset~\cite{butler12eccv-sintel} with up to 50 frames each,
containing fast articulated motion and large camera motion.

\textbf{NVIDIA.} The seven scenes of the NVIDIA dynamic scene
dataset~\cite{yoon20cvpr-nvidiadataset}---Balloon1, Balloon2,
Jumping, Playground, Skating, Truck, and Umbrella---each with 12
frames.

\textbf{DyCheck.} The seven scenes of the DyCheck iPhone
dataset~\cite{gao22neurips-dycheck}---apple, block, paper-windmill,
space-out, spin, teddy, and wheel---with 277 to 473 frames each,
captured by a handheld iPhone with realistic camera motion and
evaluated from two static test cameras.

\textbf{Syn4D.} A synthetic benchmark of eight 100-frame sequences
with ground-truth cameras and geometry, in two sub-datasets, Bear
and RoboCasa. The four \emph{Bear} sequences (bear1--bear4) place a scanned Christmas-bear
model from Artec~3D~\cite{artec_christmas_bear} in
a Replica~\cite{straub19arxiv-replica} background, each with a
different motion of the bear. The four RoboCasa sequences
(casa1--casa4) are rendered from RoboCasa~\cite{nasiriany24rss-robocasa}
scenes.

%% file: sections/app-implementation.tex
\section{Implementation Details}
\label{app:impl}

Throughout,
$\mathbf{u}_{it}\in\mathbb{R}^{2}$
is the pixel of track $i$ at frame $t$.

\subsection{Online Tracking and Query Management}
\label{app:impl-tracking}

\textbf{Online query sampling.} At every frame, candidate queries
$\mathbf{c}$ are drawn uniformly and accepted iff
\begin{equation}
\min_{\mathbf{p}\in\mathcal{R}_t}
\norm{\mathbf{c}-\mathbf{p}}>r,
\label{eq:app-resample}
\end{equation}
where $\mathcal{R}_t$ contains all tracks visible
at frame $t$ and all queries added in recent frames.

\textbf{Mask filter.} On the static side, a track whose
observations fall inside the mask more than $20\%$ of the
time leaves the static set (meaning it lives on the moving
object); otherwise only its masked observations are dropped. On
the dynamic side the masks play the opposite role.

\subsection{Observation Filtering}
\label{app:impl-filter}

\textbf{Large 2D drift (static tracks).} A static track that suddenly disagrees
with the motion of its neighbors should be excluded. For every transition $t-1\to t$ and
every track $i$ visible at both frames, define the flow and its
local reference,
\begin{equation}
\mathbf{f}_{it}=\mathbf{u}_{it}-\mathbf{u}_{i,t-1},
\qquad
\bar{\mathbf{f}}_{it}=
\operatorname*{median}_{k\in\mathcal{N}_{it}}\mathbf{f}_{kt},
\label{eq:app-flow}
\end{equation}
where $\mathcal{N}_{it}$ collects the other tracks visible at both
frames whose position lies in a $w\times w$ window centered
on $\mathbf{u}_{it}$, and the median is taken component-wise. A
transition is \emph{bad} iff the relative-motion error exceeds a
threshold,
\begin{equation}
e_{it}=\norm{\mathbf{f}_{it}-\bar{\mathbf{f}}_{it}}>\theta_t.
\label{eq:app-drift}
\end{equation}
The threshold adapts to the motion speed of the frame: with
$m_t=\operatorname*{median}_{k}\norm{\mathbf{f}_{kt}}$ the median
flow magnitude over all tracks visible at the transition,
\begin{equation}
\theta_t=\max\left(\theta_{\min},\;\kappa\,m_t\right).
\label{eq:app-drift-thresh}
\end{equation}
For every track with at least one bad transition, the visible timeline
is split at the bad transitions and only the longest contiguous
segment is kept.

\textbf{Depth filters.} Each surviving observation, static or
dynamic, is assigned the depth $d_{it}$ of its nearest depth-map
pixel when lifted to 3D. Observations with
excessive depth,
\begin{equation}
d_{it}>\tau_{z}
\qquad(\text{typically }\tau_{z}=100),
\label{eq:app-depth-max}
\end{equation}
are discarded, as monocular depth is unreliable far from the
camera. Observations on depth discontinuities, detected by
thresholding the second-order Sobel gradient magnitude of the depth
map at the sample location, are discarded as well, since the
nearest-pixel depth lookup is unreliable there.

\textbf{XM$^2$ residual trim.} After the first solve of the static
problem, the per-observation residual
\begin{equation}
r_{it}=\omega_{it}
\norm{C_t(s_t\tilde{\mathbf{u}}^{s}_{it})
+\mathbf{c}_t-\mathbf{x}^{s}_i}^{2}
\label{eq:app-xm2}
\end{equation}
is computed, observations above the $P$-th percentile of
$\{r_{it}\}$ are removed ($P=95$ in our experiments), and the
system is re-solved as described in
Section~\ref{sec:pipe-static}.

\subsection{Node Observation Term in the Dynamic Solve}
\label{app:impl-node}

Each EDG node is itself a tracked point with lifted observations
$\mathbf{y}^{d}_{jt}$ at the frames
$(j,t)\in\mathcal{E}_{\mathrm{node}}$ where it is observed. In the
implementation, the segment solves add to~\eqref{eq:dyn} a node term
that asks every node to follow its own observations,
\begin{equation}
\lambda_{\mathrm{node}}\,\mathcal{L}_{\mathrm{node}},
\qquad
\mathcal{L}_{\mathrm{node}}
=\sum_{(j,t)\in\mathcal{E}_{\mathrm{node}}}
\norm{\mathbf{g}_j+\mathbf{p}_{j,t}-\mathbf{y}^{d}_{jt}}^{2}.
\label{eq:node}
\end{equation}
Propositions~\ref{prop:dyn-rotation-only}
and~\ref{prop:dyn-sdp-bm} hold unchanged
(Appendix~\ref{app:dyn-rotation-only}).

\subsection{Fusion of Segment Solutions}
\label{app:impl-fusion}

Positions are averaged over all sequences
that solved node $j$ at frame $t$, each solution weighted by
$1/(1+\delta)$, where $\delta$ is the number of frames from $t$ to
the nearest frame in which node $j$ is directly observed within that
sequence, so that extrapolated poses count less than observed ones.
Rotations cannot be averaged directly, since each sequence fixes its
own gauge at its rest frame, but their relative motion does not
depend on that choice: every sequence $s$ that contains frames $t$
and $t'$ provides the relative rotation
$\bar R^{s}_{j,tt'}=R^{s}_{j,t'}(R^{s}_{j,t})^{\top}$. For each node
we recover all its rotations at once by rotation synchronization,
\begin{equation}
\min_{\{R_{j,t}\}\subset\mathrm{SO}(3)}\;
\sum_{s}\sum_{t,t'}\norm{R_{j,t'}-\bar R^{s}_{j,tt'}R_{j,t}}_F^2,
\label{eq:rot-sync}
\end{equation}
which we solve with a local solver, fixing the gauge by setting
$R_{j,t}=I_3$ at the first frame in which node $j$ is observed. Solving~\eqref{eq:rot-sync} with \XM is left to future work.

\subsection{Long-Range Dynamic Tracks}
\label{app:impl-long-tracks}

Raw dynamic tracks are short and, on repetitive motion, swap the
identity of similar-looking parts, so we optionally assemble them
into long-range tracks in three steps. \textbf{(a) Skeleton.} A
sparse but more reliable long-range tracker,
CoWTracker~\cite{lai26arxiv-cowtracker}, is restarted
periodically from a grid inside the object mask and run forward and
backward, so that every skeleton track spans the whole video; tracks
with large 2D jumps, inconsistent color, or low coverage are filtered
out. \textbf{(b) Attachment.} Short online
tracks~\cite{zholus25iccv-tapnext} are attached to every skeleton
point whose percentile distance to them is small. \textbf{(c)
Averaging.} Each skeleton point is replaced by the per-frame mean of
its attached tracks, giving one long track per skeleton point;
unattached online tracks are kept as additional shorter tracks.

\subsection{GPU Solver}
\label{app:impl-gpu}

\XMFD implements \textsc{Rtr-tCG} and the Riemannian staircase of
Algorithm~\ref{alg:staircase} on the GPU in double precision. The
data matrices $Q$ and $Q_h$ are obtained by eliminating the linearly
entering translations, landmarks, and node translations, an instance
of variable projection that stays efficient when sparsity is
exploited~\cite{papalia26icra-sparseavariable}. We store $Q$
densely, so each inner conjugate-gradient iteration costs one dense
matrix product for the Euclidean Hessian action $2\dot UQ$, computed
with cuBLAS~\cite{cublas129}. The tangent-space projections and the
retractions, a batched Gram--Schmidt QR for the Stiefel blocks and a
renormalization onto the sphere for the scales, are custom CUDA
kernels; the dynamic problem~\eqref{eq:dyn-bm} reuses them without
the scale variables.